\documentclass{article} %

\usepackage[preprint]{colm2025_conference}

\usepackage{microtype}
\usepackage{hyperref}
\usepackage{url}
\usepackage{booktabs}

\usepackage{lineno}

\usepackage[utf8]{inputenc} %
\usepackage[T1]{fontenc}    %
\usepackage{natbib}
\usepackage{url}            %
\usepackage{booktabs}       %
\usepackage{amsfonts}       %
\usepackage{nicefrac}       %
\usepackage{microtype}      %
\usepackage{enumitem}

\usepackage{graphicx}
\usepackage{sidecap}
\usepackage{diagbox}
\usepackage[small]{caption}
\usepackage{subcaption}
\usepackage{comment} 
\usepackage{array}
\usepackage[export]{adjustbox}

\usepackage{float} %
\usepackage{wrapfig}
\usepackage{mathtools}
\usepackage{xspace}
\usepackage{bm} %

\usepackage{booktabs}
\usepackage{multirow}
\usepackage{multicol}
\usepackage{graphicx}
\usepackage{adjustbox}

\usepackage[capitalize]{cleveref}

\definecolor{darkblue}{rgb}{0, 0, 0.5}
\hypersetup{colorlinks=true, citecolor=darkblue, linkcolor=darkblue, urlcolor=darkblue}

\newcommand{\para}[1]{\noindent \textbf{#1}}

\newif\ifhidecomments

\ifhidecomments
    \newcommand{\chenhao}[1]{}
    \newcommand{\mingxuan}[1]{}
    \newcommand{\hanchen}[1]{}
\else
    \newcommand{\chenhao}[1]{\textcolor{blue}{[\textsc{Chenhao}: #1]}}
    \newcommand{\mingxuan}[1]{\textcolor{green!70!blue}{[\textsc{Mingxuan}: #1]}}
    \newcommand{\hanchen}[1]{\textcolor{magenta!80!brown}{[\textsc{Hanchen}: #1]}}
\fi

\definecolor{humanpurple}{RGB}{235, 222, 240} 
\definecolor{mypurple}{RGB}{147,112,219} 
\definecolor{myorange}{RGB}{255,165,0} 
\definecolor{commentgray}{RGB}{86, 101, 115}
\definecolor{mygray}{RGB}{169,169,169}
\definecolor{aired}{RGB}{255,180,181}

\usepackage{listings}

\usepackage{parcolumns}
\lstdefinestyle{datalogstyle}{
        basicstyle={\tt \scriptsize},  %
	xleftmargin={6pt},
        xrightmargin={6pt},
        columns=flexible,
        breakindent=0pt,
        breaklines=true, 
	frame=tb,
	stepnumber=1,
	firstnumber=1,
	numberfirstline=true,
	tabsize=2,
	extendedchars=true,
	breaklines=true,
	columns=fullflexible,
	keepspaces=true,
	escapeinside={@}{@},
	firstnumber=last,
	captionpos=b, 
	commentstyle=\color{black!65},
	numberstyle=\tiny\color{black!65},
	stringstyle=\color{codepurple},
	breakatwhitespace=false, 
	keepspaces=true,              
        mathescape=true, 
	numbersep=5pt,                  
	showspaces=false,                
	showstringspaces=false,
	showtabs=false,
	aboveskip={0.8\baselineskip},
	belowskip={0.2\baselineskip},
}
\newcommand{\hypogenic}{\textsc{HypoGeniC}\xspace}
\newcommand{\hypoeval}{\textsc{HypoEval}\xspace}
\newcommand{\hyporefine}{\textsc{HypoRefine}\xspace}
\newcommand{\llama}{\textsc{Llama-3.3-70B-Instruct}\xspace}
\newcommand{\llamaeightb}{\textsc{Llama-3.1-8B-Instruct}\xspace}
\newcommand{\gpt}{\textsc{GPT-4o-mini}\xspace}
\newcommand{\gptsmall}{\textsc{GPT-mini}\xspace}
\newcommand{\llamasmall}{\textsc{Llama-70B}\xspace}

\definecolor{pastelgreen}{RGB}{50,205,50}
\newcommand{\increase}{\textcolor{pastelgreen}{\bm{$\uparrow$} }}
\newcommand{\decrease}{\textcolor{red}{\bm{$\downarrow$} }}
\newcommand{\nochange}{\textcolor{gray}{\bm{$=$} }}

\title{HypoEval: Hypothesis-Guided Evaluation for Natural Language Generation}

\author{Mingxuan Li, Hanchen Li, Chenhao Tan\\
Department of Computer Science\\
University of Chicago\\
Chicago, IL 60637, USA\\
\texttt{\{mingxuanl, lihanc2002, chenhao\}@uchicago.edu}
}

\begin{document}

\maketitle

\renewcommand{\thefootnote}{\arabic{footnote}}
\footnotetext[1]{Code available at \url{https://github.com/ChicagoHAI/HypoEval-Gen}. We also provide off-the-shelf 0-shot evaluators for summarization and story generation at \url{https://github.com/ChicagoHAI/HypoEval}.}

\begin{abstract}

Large language models (LLMs) have demonstrated great potential for automating the evaluation of natural language generation. Previous frameworks of LLM-as-a-judge fall short in two ways: they either use zero-shot setting without consulting any human input, which leads to low alignment, or fine-tune LLMs on labeled data, which requires a non-trivial number of samples.
Moreover, previous methods often provide little reasoning behind automated evaluations. 
In this paper, we propose \hypoeval, {\bf Hypo}thesis-guided {\bf Eval}uation framework, which first uses a small corpus of human evaluations to generate more detailed rubrics for human judgments
and then incorporates a checklist-like approach to combine LLM's assigned scores on each decomposed dimension to acquire overall scores. With only 30 human evaluations, HypoEval achieves state-of-the-art performance 
in alignment with both human rankings (Spearman correlation) and human scores (Pearson correlation), on average outperforming G-Eval by 11.86\% and fine-tuned \llamaeightb with at least 3 times more human evaluations by 11.95\%.
Furthermore, we conduct systematic studies to assess the robustness of \hypoeval, highlighting its effectiveness as a reliable and interpretable automated evaluation framework.\textsuperscript{1}

\end{abstract}

\section{Introduction}
\label{sec:introduction_02}

Automated evaluation of natural language generation has been an important and challenging task with the rapid development of automated systems for summarization, translation, open-ended story generation, and more \citep{fang2024multillmtextsummarization, yao2024benchmarkingmachinetranslationcultural}. Traditional lexical metrics such as BLEU \citep{papineni2002bleu} and ROUGE \citep{lin2004rouge} have been shown to have low agreement with human judgments \citep{krishna-etal-2021-hurdles}. With the advancements of large language models (LLMs), recent research has extensively focused on LLM-as-a-judge, or using LLMs to perform reference-free automated evaluations of natural language generation \citep{chen2023exploringuselargelanguage,fu2023gptscoreevaluatedesire,gu2025surveyllmasajudge}.

Following \citet{li2025generationjudgmentopportunitieschallenges}, we broadly categorize existing LLM-based automated evaluation frameworks into prompting-based and tuning-based methods. On one hand, prompting-based evaluation methods such as G-Eval \citep{liu2023gevalnlgevaluationusing} mostly use a zero-shot approach, which imposes a strict yet unnecessary restriction on the use of human evaluations or groundings. As a result, they lead to limited correlations with human annotations and leave room for improvement \citep{bavaresco2024llmsinsteadhumanjudges,krumdick2025freelabelslimitationsllmasajudge}. On the other hand, tuning-based methods \citep{yue2023automaticevaluationattributionlarge, liu-etal-2024-x} require a large corpus of high-quality training data, with performance constrained to the specific dataset that they are trained on \citep{liu2025aligninghumanjudgementrole}, and can be computationally expensive or hard to apply to proprietary models. Furthermore, both categories of methods often lack explainability in their evaluation process. Although recent works on checklist-based frameworks for evaluating instruction-following or factuality \citep{que2024hellobenchevaluatinglongtext,min2023factscorefinegrainedatomicevaluation,tan2024proxyqaalternativeframeworkevaluating} shed light on providing reasoning behind evaluations, their performance on evaluating other aspects of text generation can be inferior to non-checklist frameworks \citep{lee2024checkevalrobustevaluationframework}. This can be due to the fundamental difficulty of decomposing subjective aspects (e.g., engagement) of texts into atomic and easy-to-verify checklists.

To address these limitations, we propose \hypoeval, the first LLM-based evaluation framework that combines state-of-the-art hypothesis generation techniques to guide judge LLMs and improve human alignment. \hypoeval consists of a light training stage for hypothesis generation and then uses hypotheses to provide evaluation scores. With a small corpus of human evaluation scores (in our implementation, we limit the number to 30), \hypoeval first uses a hypothesis generation framework to generate high-quality hypotheses, which are framed as decomposed dimensions for evaluation,  from human evaluation results and existing literature on evaluation. The decomposed dimensions serve as rubrics and break down a subjective aspect of evaluation into different attributes that are easier for an LLM to understand. Then in the  evaluation stage, with each decomposed dimension formulated as a non-binary checklist (i.e., the answer to the checklist can be a range of numbers on the Likert scale), \hypoeval combines an evaluator LLM's assigned scores on each decomposed dimension and gets an overall score for an evaluated text. We show that with only small-scale human evaluations and $O(N)$ computational complexity, \hypoeval is able to achieve state-of-the-art performance on representative tasks: on average outperforming G-Eval by 11.86\% and fine-tuned \llamaeightb - with more than 3 times more human evaluation scores - by 11.95\%.

\begin{figure}[t]
    \centering
    \makebox[\textwidth]{
        \includegraphics[width=1\textwidth]{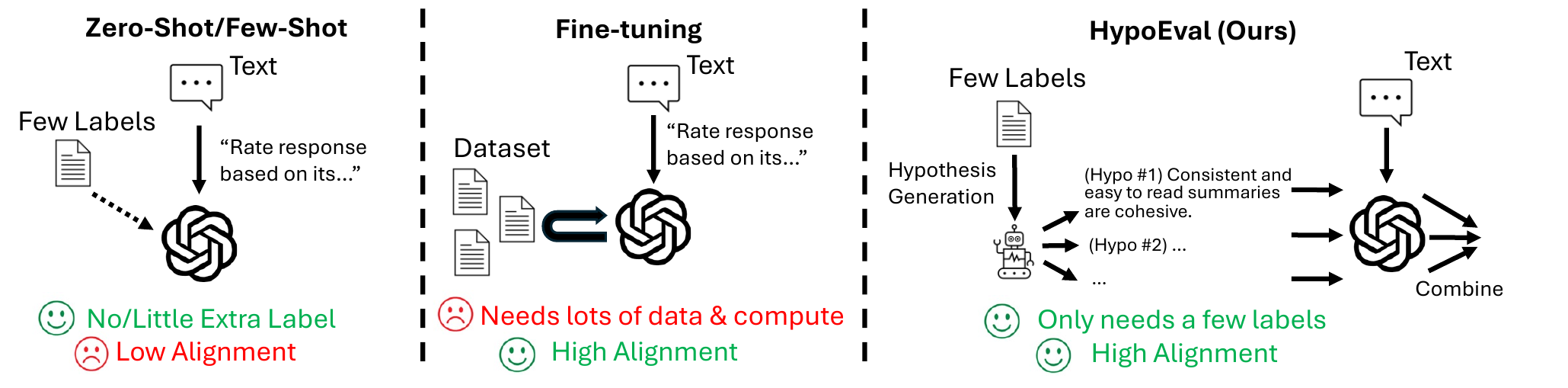}
    }
    \caption{A Comparison between previous methods and HypoEval. We achieve high-alignment and explainable evaluation with only a few human labels per dataset.}
    \label{fig:example}
\end{figure}

To summarize, our main contributions are as follows:

\begin{itemize}[leftmargin=*,itemsep=0pt,topsep=-2pt]
    \item We introduce \hypoeval, a tuning-free, sample-efficient framework for evaluating natural language generation.
    \item We demonstrate that \hypoeval achieves state-of-the-art performance in terms of correlation with human judgments across multiple datasets.
    \item With the generated hypotheses or decomposed dimensions in HypoEval, we provide more interpretable explanations in automated evaluations than previous methods.
\end{itemize}

\section{Methods}
\label{sec:methods}
We first give a formulation of hypothesis-guided text evaluation.  In direct scoring, we evaluate on an input-output pair $(x,y)$, where $x$ is the prompt for generation (e.g. source text for summarization), and $y$ is the generated content (e.g. summary). With an LLM $\mathcal{M}$ and an instruction prompt $I$ that consists of task descriptions and definition of the evaluated aspect (e.g. coherence of summaries), we want to produce a score ${\mathcal{M}}(x,y,I)$ that matches human score $s$ well (usually measured in Pearson or Spearman correlation). 

In hypothesis-guided text evaluation, we first generate a hypothesis bank $\mathcal{H} = \{h_1, h_2,\dots, h_n\}$ from a training set $S_{\text{tr}} = \{(x_1, y_1, s_1), \dots, (x_m, y_m, s_m)\}$ and a corpus of summaries $\mathcal{L}$ of relevant literature, where $x_i$ and $y_i$ are inputs and outputs, and $s_i$ are scores given by human experts on $(x_i, y_i)$. Here, each hypothesis is formulated as a rubric on a decomposed dimension. Ideally, the hypothesis bank $\mathcal{H}$ contains multiple decomposed dimensions that human experts consider when giving the scores in $\mathcal{S}_{\text{tr}}$. Then, we use a checklist-like approach to use the hypotheses to evaluate on unseen sample $(x,y)$ and give a score ${\mathcal{M}}(x,y,I, \mathcal{H})$.

\paragraph{Hypothesis Generation from Small-scale Data and Literature}
\label{sec:hypothesis-generation}

Following \hypogenic \citep{Zhou_2024} and \hyporefine \citep{liu2025literaturemeetsdatasynergistic}, we utilize both data-driven and literature-driven hypothesis generation approaches to generate the hypothesis bank $\mathcal{H}$. That is, with an LLM $\mathcal{M}$ and hypothesis generation algorithm $g$, we have $\mathcal{H} = g_{\mathcal{M}}(S_{\text{tr}}, \mathcal{L})$.

Specifically, we build upon \hyporefine and introduce extensions tailored for continuous value prediction. Given a small-scale training set $S_{\text{tr}} = \{(x_1, y_1, s_1), \dots, (x_m, y_m, s_m)\}$ (for main experiments, we set $|S_{\text{tr}}| = 30$) and the corpus of summaries of relevant literature $\mathcal{L}$ (details of literature collection and summary generation in \cref{appendix:implementation_details:hypgen}), we first want to generate a hypothesis bank $\mathcal{H}$. During the initial stage, an LLM-based hypothesis generation agent $\mathcal{M}_G$ is prompted with a set of initial data points $S_{\text{init}}\subset S_{\text{tr}}$ and $\mathcal{L}$ to generate an initial hypothesis bank $\mathcal{H}^{\text{init}} = \mathcal{M}_G(S_{\text{init}}, \mathcal{L})$, and set $\mathcal{H}^0 = \mathcal{H}^{\text{init}}$. Inspired by the Upper Confidence Bound algorithm for strategic regression \citep{NIPS2016_d79aac07}, for each $h\in \mathcal{H}^0$, we use an evaluator $\mathcal{M}_E$ to evaluate on all $(x_i, y_i)$ in $S_{\text{init}}$ based on the detailed rubric stated in $h$ and give a score ${\mathcal{M}_E}(x_i,y_i, I, h)$, where $I$ is the instruction prompt. Then, we set the reward for $h$ by:

\begin{equation*}
    r_h \coloneqq \frac{1}{|S_{\text{init}}|} \sum_{(x_j, y_j, s_j)\in S_{\text{init}}} (a - b (s_j - {\mathcal{M}_E}(x_j,y_j, I, h))^2) + \alpha\sqrt{\frac{\log|S_{\text{init}}|}{|S_{\text{init}}|}}
\end{equation*}

where $\alpha$ is the reward coefficient that controls the exploration term of the reward function, and $a$, $b$ are coefficients that control the range of the exploitation term.

In the update stage, we iterate over all data points in $S_{\text{update}} = S_{\text{tr}}\setminus S_{\text{init}}$. For time $t$, we consider the training sample $(x_t, y_t, s_t)\in S_{\text{update}}$. We first choose the top $k$ hypotheses $\mathcal{H}_{\text{top}}$ with the highest reward from $\mathcal{H}^{t-1}$. Then for each $h\in \mathcal{H}_{\text{top}}$, we update the reward with:

\begin{equation*}
    r_h \coloneqq \frac{1}{|S_{h}^t|} \sum_{(x_j, y_j, s_j)\in S_{h}^t} (a - b(s_j - {\mathcal{M}_E}(x_j,y_j, I, h))^2 ) + \alpha\sqrt{\frac{\log (t+ |S_{\text{init}}|)}{|S_h^t|}}
\end{equation*}

where $S_{h}^t$ is the set of training samples seen by hypothesis $h$ at time $t$.

For all hypotheses from $\mathcal{H}_{\text{top}}$, if at least $w_{\text{hyp}}$ predicted a score with $|{\mathcal{M}_E}(x_t,y_t, I, h) - s_t| > \theta$, where $\theta$ is the threshold for identifying a wrong prediction, the datapoint $(x_t, y_t, s_t)$ is added to a wrong sample bank $\mathcal{W}$. Once $|\mathcal{W}|\geq w_{\text{max}}$, a new set of hypotheses $\mathcal{H}_{\mathcal{W}}$ is generated using $\mathcal{W}$ and $\mathcal{L}$ by an iterative refinement process:

\begin{equation*}
\begin{aligned}
    \mathcal{H}_{\mathcal{W}}^0 & = \mathcal{M}_G(\mathcal{W}), \\
    \mathcal{H}_{\mathcal{W}}^i, i>0 & = \begin{cases}
        \mathcal{M}_R(\mathcal{H}_{\mathcal{W}}^{i-1}, \mathcal{L}) & \text{if } i \bmod 2 = 0 \\ \mathcal{M}_R(\mathcal{H}_{\mathcal{W}}^{i-1}, \mathcal{W})& \text{if } i \bmod 2 = 1.
    \end{cases}
\end{aligned}
\end{equation*}

The refinement finishes in $N_{\operatorname{refine}}$ rounds, and we get $\mathcal{H}_{\mathcal{W}} = \mathcal{H}_{\mathcal{W}}^{N_{\operatorname{refine}}}$. The wrong sample bank $\mathcal{W}$ is set to $\emptyset$ afterwards. For $\mathcal{H}^t$, we choose $H_{\text{max}}$ hypotheses with the highest reward from $\mathcal{H}_{\text{top}}\cup \mathcal{H}_{\mathcal{W}}$. 

Following \cite{liu2025literaturemeetsdatasynergistic}, to accommodate for that literature-based hypotheses can be undervalued during the update stage, we use a union approach to combine hypotheses from literature only $\mathcal{H}_{\mathcal{L}} = \mathcal{M}_G(\mathcal{L})$ and $\mathcal{H}^{|S_{\text{update}}|}$. Specifically, for a final hypothesis bank with size $H_{\text{max}}$, we first remove redundant hypotheses from $\mathcal{H}_{\mathcal{L}} = \mathcal{M}_G(\mathcal{L})$ and $\mathcal{H}^{|S_{\text{update}}|}$, and then randomly choose at most $\frac{H_{\text{max}}}{2}$ from each of them for the final hypothesis bank $\mathcal{H}$.

\paragraph{Hypothesis selection.}
\label{sec:hypothesis-selection}

Since that we are encouraging diverse and novel hypotheses in the hypothesis generation process by both the exploration term in reward and the incorporation of information from literature, it is possible that we also include some hypotheses that are interesting but are not suitable for a specific evaluation task. To accommodate this, we perform hypothesis selection from $\mathcal{H}$ based on the hypotheses' performance on $S_{\text{tr}}$.

Specifically, we choose the top $H_{\text{eval}}$ hypotheses with the highest Pearson correlations with human scores on $S_{\text{tr}}$:

\begin{equation*}
    \mathcal{H}_{\text{eval}} = \arg \max_{\mathcal{H}'\subset \mathcal{H}, |\mathcal{H}'| = H_{\text{eval}}} \sum_{h\in \mathcal{H}'}r(h, S_{\text{tr}}),
\end{equation*}
where $r(h, S_{\text{tr}})$ is the Pearson correlation between the human scores and the scores given by the evaluator agent $\mathcal{M}_E$ based specifically on the decomposed dimension stated in $h$.

\paragraph{Hypothesis-guided text evaluation}
\label{sec:hypothesis-based-evaluation}

For each hypothesis $h$, we first evaluate the text based solely on the dimension entailed in $h$ (e.g. logical structure of events for evaluating coherence of summaries) with chain-of-thought prompting \citep{NEURIPS2022_9d560961}. The evaluator $\mathcal{M}_E$ is asked to give a score rating between 1 and 5. Then, since different hypotheses in $\mathcal{H}_{\text{eval}}$ often entail different decomposed dimensions that are important for a holistic evaluation, we combine the scores on all hypotheses to acquire the final overall score for a sample:

\begin{equation*}
    \mathcal{M}_E (x,y,I,\mathcal{H}_{\text{eval}}) = \frac{1}{|\mathcal{H}_{\text{eval}}|}\sum_{h\in\mathcal{H}_{\text{eval}}}\mathcal{M}_E(x,y,I,h).
\end{equation*}

For more detailed information of the implementation, please refer to \cref{appendix:implementation_details:hypgen}.

\paragraph{Efficiency Analysis} 
The computational complexity of \hypoeval can be separated into two parts: a preparation stage of hypothesis generation and selection, and an evaluation stage of hypothesis-guided automated evaluation. Let $N$ be the total number of texts to be evaluated. For the preparation stage, the complexity of hypothesis generation can be expressed as $O(N_{\text{paper}}+(k+N_{\text{refine}})|S_{\text{tr}}| + H_{\text{max}}^2)$, where $N_{\text{paper}}$ is the number of papers as relevant literature, and the complexity of hypothesis selection is $O(H_{\text{max}}|S_{\text{tr}}|)$. For the evaluation stage, computational complexity is $O(H_{\text{eval}} N)$. Under our setting where $S_{\text{tr}}, k,N_{\text{refine}}, N_{\text{paper}}, H_{\text{max}} \leq 30$ and $H_{\text{eval}}=5$, the total complexity of \hypoeval can be expressed as $O(N)$ and is equivalent to other pointwise evaluators.

\section{Experiment Setup}
\label{sec:experiments}

To demonstrate the effectiveness of our hypothesis-guided evaluation framework, we first compare our method with baselines on two NLG tasks with four datasets. We report both Spearman correlation and Pearson correlation to account for both alignment with human rankings and with human scores. 

\paragraph{Tasks and datasets.} We report two representative tasks, summarization and open-ended story generation, to evaluate our framework.

For the summarization task, we choose SummEval \citep{fabbri2021summevalreevaluatingsummarizationevaluation} and NewsRoom \citep{grusky2020newsroomdataset13million} for our experiments. SummEval consists of 100 source texts, each with 16 summaries generated by different models, and is annotated on four aspects: coherence (CH), consistency (CON), fluency (FLU), and relevance (RE). We use the average of annotation scores of 3 human experts. NewsRoom has 60 source texts and 7 summaries for each text, and is annotated on four aspects: coherence (CH), informativeness (INF), fluency (FLU), and relevance (RE). For each dataset, we randomly sample 30 source text-summary pairs and their human evaluation scores as training data, and perform automated evaluation on summaries of 40 source texts for SummEval and summaries of 30 source texts for NewsRoom, with a total of 640 and 210 summaries, respectively. Following \citep{liu2023gevalnlgevaluationusing}, we report summary-level Spearman correlations and Pearson correlations.

For the open-ended story generation task, we use HANNA \citep{chhun2022humancriteriaautomaticmetrics} and part of WritingPrompt (WritingPrompt-A) with human annotations collected by \cite{chiang2023largelanguagemodelsalternative}. HANNA includes 96 writing prompts, each with 11 stories annotated on 6 aspects: coherence (CH), complexity (CX), empathy (EM), engagement (EG), relevance (RE), and surprise (SU). WritingPrompt-A consists of 400 prompt-story pairs, each annotated on grammaticality (GRA), cohesiveness (COH), likability (LIK), and relevance (RE). For both datasets, we choose 30 prompt-story pairs for training. For HANNA, we randomly select 60 prompts, each with 11 stories, for testing, and report story-level Spearman correlations and Pearson correlations. For WritingPrompt-A, we choose 300 prompt-story pairs for testing. Due to the lack of story batches grouped by the same prompts, we report dataset-level Spearman correlations and Pearson correlations.

\paragraph{Baselines and implementation.} \label{sec:experiments:baselines} We largely characterize baselines into two categories: zero-shot evaluators that do not consult human evaluations, and data-augmented evaluators that utilize specific datasets or are trained on specific tasks. 

For zero-shot evaluators, we include ROUGE-L \citep{lin2004rouge}, BERTScore \citep{zhang2020bertscoreevaluatingtextgeneration}, 
G-Eval \citep{liu2023gevalnlgevaluationusing} with probabilities and automatically generated chain-of-thought (CoT) prompting, and \textit{direct scoring} (evaluator LLM assigns a score directly to a text) with CoT. We also consider PairS-beam 
\citep{liu2025aligninghumanjudgementrole}, a pairwise ranking evaluator, for comparison in Spearman correlation. For comparison with other checklist-based approaches, we implemented CheckEval, \citep{lee2024checkevalrobustevaluationframework} on own, because human-curated key components for each aspect were used in the original paper but not released. We leverage the LLM's prior knowledge to generate atomic checklists.

For data-augmented evaluators, we include UniEval \citep{zhong2022unifiedmultidimensionalevaluatortext} and BARTScore \citep{yuan2021bartscoreevaluatinggeneratedtext}, which are task-specific evaluators trained with large corpora of data. We also fine-tune \llamaeightb on each aspect of each dataset with 30 (FT-A) and 200 (FT-B) human-annotated data points. Due to the use of significantly larger amount of training data than \hypoeval, we regard UniEval, BARTScore, and FT-B as strong but not directly comparable methods. Furthermore, we consider direct scoring with few-shot demonstrations from human-annotated data. For the WritingPrompt-A dataset, due to its size (see \cref{sec:experiments:baselines}) and the lack of references, we implement FT-B with 100 human evaluations and omit the performance of reference-based evaluators.
Implementation details of baselines are available in \cref{appendix:implementation_details:baselines}.

Our framework works for any LLM $\mathcal{M}$. In the experiments, we utilize two models, \gpt \citep{openai2023gpt4} and \llama \citep{dubey2024llama3herdmodels} to reflect a range of different model sizes. We abbreviate \gpt as \textsc{GPT-mini} and \llama as \textsc{Llama-70B}. For main experiments, we let $H_{\text{eval}}=5$, and the evaluator model $\mathcal{M}_E$ is the same as the hypothesis generator model $\mathcal{M}_G$.

\begin{table*}[t]
    \centering
    \centering
\resizebox{1\textwidth}{!}{%
\begin{tabular}{@{}llcccccccccccccccc@{}}

\toprule

 &  & \multicolumn{8}{c}{SummEval} & \multicolumn{8}{c}{NewsRoom} \\

\cmidrule(lr){3-10} \cmidrule(lr){11-18}

Models & Methods  & \multicolumn{2}{c}{CH} & \multicolumn{2}{c}{CON} & \multicolumn{2}{c}{FLU} & \multicolumn{2}{c}{RE} & \multicolumn{2}{c}{CH} & \multicolumn{2}{c}{INF} & \multicolumn{2}{c}{FLU} & \multicolumn{2}{c}{RE}\\

\cmidrule(lr){3-4} \cmidrule(lr){5-6} \cmidrule(lr){7-8} \cmidrule(lr){9-10} \cmidrule(lr){11-12} \cmidrule(lr){13-14} \cmidrule(lr){15-16} \cmidrule(lr){17-18}

& & $\rho$ & $r$ & $\rho$ & $r$ & $\rho$ & $r$ & $\rho$ & $r$ & $\rho$ & $r$ & $\rho$ & $r$ & $\rho$ & $r$ & $\rho$ & $r$ \\
\midrule
\multirow{6}{*}{Other} 
    & ROUGE-L & 0.10  & 0.12  & 0.14  & 0.18  & 0.09  & 0.12  & 0.19  & 0.20  & 0.08  & -0.11  & 0.08  & -0.08  & 0.07  & -0.10  & 0.08  & -0.06  \\
    & BERTScore & 0.26  & 0.27  & 0.21  & 0.21  & 0.18  & 0.17  & 0.38  & 0.40  & 0.31  & 0.18  & 0.31  & 0.20  & 0.33  & 0.17  & 0.28  & 0.18  \\
    & BARTScore & 0.47  & 0.50  & 0.26  & 0.25  & 0.25  & 0.25  & 0.35  & 0.35  & 0.66  & 0.72  & 0.59  & 0.75  & 0.64  & 0.70  & 0.56  & 0.74  \\
    & UniEval & 0.56  & 0.51  & 0.47  & 0.63  & 0.39  & 0.47  & 0.43  & 0.42  & --  & --  & --  & --  & --  & --  & --  & --  \\
    & FT-A & 0.47  & 0.48  & 0.38  & 0.40  & 0.31  & 0.39  & 0.40  & 0.40  & 0.50  & 0.51  & 0.53  & 0.56  & 0.58  & 0.59  & 0.48  & 0.56  \\
    & FT-B & 0.57  & 0.59  & 0.57  & 0.61  & 0.46  & 0.57  & 0.46  & 0.47  & 0.61  & 0.62  & 0.69  & 0.72  & 0.61  & 0.63  & 0.61  & 0.71  \\

\midrule
\multirow{6}{*}{\gptsmall} \\
    & direct scoring & 0.50 & 0.49 & \bf 0.54 & 0.62 & 0.22 & 0.23 & 0.47 & 0.51 & 0.51 & 0.51 & 0.47 & 0.46 & 0.59 & 0.59 & 0.46 & 0.44 \\
    & few-shot scoring & 0.37 & 0.38 & 0.47 & 0.49 & 0.33 & 0.35 & 0.42 & 0.44 & 0.60 & 0.61 & 0.57 & 0.62 & 0.65 & 0.67 & 0.53 & 0.60 \\
    & G-Eval & 0.54 & 0.54 & 0.51 & \bf 0.65 & 0.31 & 0.30 & 0.47 & 0.55 & 0.59 & 0.53 & 0.58 & 0.52 & 0.63 & 0.62 & 0.51 & 0.44 \\
    & PairS-beam & 0.52 & -- & 0.53 & -- & 0.31 & -- & 0.49 & -- & 0.53 & -- & 0.61 & -- & 0.43 & -- & 0.55 & -- \\
    & CheckEval & 0.47 & 0.47 & 0.34 & 0.36 & 0.39 & 0.38 & 0.39 & 0.43 & 0.53 & 0.56 & 0.34 & 0.42 & 0.57 & 0.57 & 0.46 & 0.52 \\
    & \hypoeval & \bf 0.58 & \bf 0.58 & 0.51 & 0.63 & \bf 0.40 & \bf 0.45 & \bf 0.54 & \bf 0.58 & \bf 0.64 & \bf 0.69 & \bf 0.62 & \bf 0.75 & \bf 0.67 & \bf 0.69 & \bf 0.60 & \bf 0.78 \\

\midrule
\multirow{6}{*}{\llamasmall} \\
    & direct scoring & 0.56  & 0.57  & 0.56  & 0.64  & 0.33  & 0.34  & 0.47  & 0.50  & 0.51  & 0.48  & 0.46  & 0.51  & 0.58  & 0.53  & 0.43  & 0.48  \\
    & few-shot scoring & 0.45  & 0.45  & \bf 0.61  & 0.67  & 0.40  & 0.47  & 0.47  & 0.48  & 0.59  & 0.60 & 0.63  & 0.68  & 0.60  & 0.62  & 0.51  & 0.69  \\
    & G-Eval &  0.61 &  0.57 & 0.51  & \bf 0.68  & \bf 0.41  & \bf 0.48  & 0.52  &  0.49 & 0.53  & 0.57  & 0.57  & 0.59  & 0.54  & 0.55  & 0.51  & 0.65  \\
    & PairS-beam &  0.60 &  -- & 0.54  & --  & 0.37  & --  & 0.50  & --  & 0.58  & --  & \bf 0.65  & --  & 0.58  & --  & \bf 0.59  & --  \\
    & CheckEval & 0.56  & 0.58  & 0.43  & 0.45  & 0.45  & 0.45  & 0.47  & 0.50  & 0.40  & 0.42  & 0.32  & 0.34  & 0.30  & 0.30  & 0.48  & 0.59  \\
    & \hypoeval & \bf 0.63  & \bf 0.63  & 0.49  & 0.62  & 0.35  & 0.35  &  \bf 0.54 &  \bf 0.56 & \bf 0.62  & \bf 0.65  & \bf 0.65  & \bf 0.74  & \bf 0.65  & \bf 0.64  & 0.52  & \bf 0.73  \\
\bottomrule
\end{tabular}
}
\label{tab:main_01}

    \bigskip
    \centering
\resizebox{1\textwidth}{!}{%
\begin{tabular}{@{}llcccccccccccccccccccc@{}}

\toprule

 &  & \multicolumn{12}{c}{HANNA} & \multicolumn{8}{c}{WritingPrompt-A} \\

\cmidrule(lr){3-14} \cmidrule(lr){15-22}

Models & Methods  & \multicolumn{2}{c}{CH} & \multicolumn{2}{c}{CX} & \multicolumn{2}{c}{EM} & \multicolumn{2}{c}{EG} & \multicolumn{2}{c}{RE} & \multicolumn{2}{c}{SU} & \multicolumn{2}{c}{GRA} & \multicolumn{2}{c}{COH} & \multicolumn{2}{c}{LIK} & \multicolumn{2}{c}{RE}\\

\cmidrule(lr){3-4} \cmidrule(lr){5-6} \cmidrule(lr){7-8} \cmidrule(lr){9-10} \cmidrule(lr){11-12} \cmidrule(lr){13-14} \cmidrule(lr){15-16} \cmidrule(lr){17-18} \cmidrule(lr){19-20} \cmidrule(lr){21-22}

& & $\rho$ & $r$ & $\rho$ & $r$ & $\rho$ & $r$ & $\rho$ & $r$ & $\rho$ & $r$ & $\rho$ & $r$ & $\rho$ & $r$ & $\rho$ & $r$ & $\rho$ & $r$ & $\rho$ & $r$\\
\midrule
\multirow{6}{*}{Other} 
    & ROUGE-L & 0.17 & 0.23 & 0.26 & 0.29 & 0.15 & 0.16 & 0.21 & 0.24 & 0.10 & 0.09 & 0.16 & 0.17 & -- & -- & -- & -- & -- & -- & -- & -- \\
    & BERTScore & 0.30 & 0.36 & 0.42 & 0.48 & 0.28 & 0.31 & 0.34 & 0.39 & 0.17 & 0.18 & 0.26 & 0.26 & -- & -- & -- & -- & -- & -- & -- & -- \\
    & BARTScore & -- & -- & -- & -- & -- & -- & -- & -- & -- & -- & -- & -- & -- & -- & -- & -- & -- & -- & -- & -- \\
    & UniEval & -- & -- & -- & -- & -- & -- & -- & -- & -- & -- & -- & -- & -- & -- & -- & -- & -- & -- & -- & -- \\
    & FT-A & 0.40 & 0.46 & 0.37 & 0.42 & 0.32 & 0.37 & 0.35 & 0.41 & 0.43 & 0.48 & 0.17 & 0.20 & 0.29 & 0.28 & 0.34 & 0.30 & 0.22 & 0.22 & 0.53 & 0.52 \\
    & FT-B & 0.40 & 0.45 & 0.52 & 0.58 & 0.38 & 0.43 & 0.44 & 0.51 & 0.46 & 0.52 & 0.33 & 0.39 & 0.37 & 0.36 & 0.51 & 0.49 & 0.36 & 0.36 & 0.59 & 0.58 \\

\midrule
\multirow{6}{*}{\gptsmall} \\
    & direct scoring & 0.47 & 0.59 & 0.51 & 0.54 & 0.35 & 0.41 & 0.49 & 0.56 & 0.48 & 0.59 & 0.37 & 0.45 & 0.47 & 0.40 & 0.56 & 0.51 & 0.41 & 0.38 & 0.63 & 0.63 \\
    & few-shot scoring & 0.47 & 0.55 & 0.44 & 0.47 & 0.40 & 0.47 & 0.46 & 0.51 & 0.42 & 0.51 & 0.36 & 0.42 & 0.40 & 0.41 & 0.55 & 0.54 & 0.42 & 0.39 & 0.65 & 0.65 \\
    & G-Eval & 0.48 & 0.62 & 0.53 & 0.56 & 0.41 & 0.47 & 0.46 & 0.55 & 0.49 & 0.61  & \bf 0.38 & \bf 0.48 & 0.53 & 0.48 & 0.58 & 0.55 & 0.43 & 0.40 & 0.66 & 0.66 \\
    & PairS-beam & 0.39 & -- & 0.51 & -- & 0.43 & -- & 0.48 & -- & 0.39 & -- & \bf 0.38 & -- & 0.20* & -- & 0.57 & -- & 0.48 & -- & 0.09* & -- \\
    & CheckEval & 0.46 & 0.55 & 0.40 & 0.41 & 0.37 & 0.40 & 0.46 & 0.50 & \bf 0.52 & 0.57 & 0.26 & 0.27 & 0.17 & 0.17 & 0.58 & 0.59 & 0.29 & 0.30 & 0.67 & 0.67 \\
    & \hypoeval & \bf 0.55 & \bf 0.67 & \bf 0.55 & \bf 0.61  & \bf 0.50 & \bf 0.57 & \bf 0.54 & \bf 0.63 & 0.49 & \bf 0.62 & \bf 0.38 & 0.45 & \bf 0.54 & \bf 0.53 & \bf 0.64 & \bf 0.60 & \bf 0.53 & \bf 0.52 & \bf 0.70 & \bf 0.68 \\

\midrule
\multirow{6}{*}{\llamasmall} \\
    & direct scoring & 0.49 & 0.59 & 0.45 & 0.46 & 0.40 & 0.46 & 0.42 & 0.48 & 0.46 & 0.55 & 0.30 & 0.33 & 0.45 & 0.44 & 0.58 & 0.58 & 0.27 & 0.24 & 0.63 & 0.63 \\
    & few-shot scoring & 0.52 & 0.62 & 0.54 & 0.56 & 0.43 & 0.46 & 0.48 & 0.51 & 0.44 & 0.50 & 0.35 & 0.35 & 0.42 & 0.40 & 0.59 & 0.57 & 0.37 & 0.36 & 0.62 & 0.61 \\
    & G-Eval & 0.53 & 0.65 & 0.49 & 0.53 & 0.37 & 0.44 & 0.45 & 0.52 & 0.48 & 0.57 & 0.37 & 0.44 & \bf 0.47 & \bf 0.45 & 0.61 & 0.61 & 0.24 & 0.19 & 0.65 & 0.63 \\
    & PairS-beam & 0.47 & -- & 0.55 & -- & 0.46 & -- & 0.46 & -- & 0.46 & -- & 0.42 & -- & 0.26* & -- & 0.57 & -- & 0.49 & -- & 0.04* & -- \\
    & CheckEval & 0.38 & 0.51 & 0.40 & 0.39 & 0.45 & 0.48 & 0.38 & 0.47 & 0.48 & 0.57 & 0.24 & 0.23 & 0.41 & 0.37 & 0.63 & 0.62 & 0.50 & 0.48 & 0.66 & 0.66 \\
    & \hypoeval & \bf 0.54 & \bf 0.67 & \bf 0.56 & \bf 0.66 & \bf 0.47 & \bf 0.54 & \bf 0.52 & \bf 0.60 & \bf 0.51 & \bf 0.63 & 0.40 & \bf 0.50 & 0.44 & 0.41 & \bf 0.63 & \bf 0.62 & \bf 0.53 & \bf 0.51 & \bf 0.69 & \bf 0.68 \\
\bottomrule
\end{tabular}
}
\label{tab:main_02}

    \caption{Evaluation results of \gpt and \llama. We report Spearman correlation ($\rho$) and Pearson correlation ($r$) for a total of 18 aspects of the 4 datasets. Some especially lower performance of PairS-beam marked with * is due to that the model frequently failed to generate pairwise preferences.}
    \label{tab:combined_main}
\end{table*}

\section{Results}
\label{sec:results_final}

\cref{tab:combined_main} presents the main results across a total of 18 aspect-dataset settings. Comparing with baselines without large-scale tuning, \hypoeval with \gpt achieves state-of-the-art (SOTA) performance on 15 settings with Spearman correlation and 16 settings with Pearson correlation, on average outperforming G-Eval with CoT by 9.8\% and 15.7\% respectively; \hypoeval with \llama achieves SOTA on 13 settings for Spearman correlation and 15 settings for Pearson correlation, outperforming G-Eval by 9.9\% and 11.8\% respectively. 

Some exceptions, such as the consistency and fluency aspects of SummEval, could be due to the human scores being highly skewed towards 5, illustrated in \cref{appendix:additional_illustrations}.

In addition, though \hypoeval is not explicitly optimized for ranking or pairwise comparison, it still outperforms the ranking-based evaluator PairS-beam on 16/18 and 13/18 settings for \gpt and \llama respectively. 

Comparing with tuning-based evaluators that use at least more than 3 times more annotated data (FT-B, BARTScore, UniEval), \hypoeval still demonstrates strong performance. For the story generation task, \hypoeval with \gpt or \llama outperforms FT-B across all settings. For the summarization task, \hypoeval on average outperforms BARTScore by 18.66\%; after excluding the exceptions of the consistency and fluency aspects of SummEval, \hypoeval on average outperforms FT-B by 4.8\%.

To further illustrate our method, we include examples of hypotheses in \cref{tab:example_hypotheses}. As demonstrated by the coherence aspect of SummEval in the table, these different hypotheses cover different decomposed dimensions of what a human would consider. For example, the first hypothesis covers that the answer should be "logically organized, with a clear introduction, body, and conclusion"; the second hypothesis highlights the "consistent tone and style", and the third one contains that the answer should not introduce "unrelated themes or topics". We include full versions of more examples in \cref{appendix:example_hypotheses}.

\begin{table*}[t]
\centering
\small
\resizebox{1\textwidth}{!}{%
\begin{tabular}{@{}p{\textwidth}@{}}
\toprule

\textbf{Example Hypotheses (Decomposed Dimensions) on SummEval - CH} \\
\midrule
- The overall structure and organization of the summary play a vital role in determining coherence scores. Summaries that are logically organized, with a clear introduction, body, and conclusion, will score higher (4 or 5), while those \dots \\
- Summaries that maintain a consistent tone and style throughout will be rated higher for coherence (4 or 5), as this consistency aids in reader comprehension. In contrast, \dots \\
-  The thematic consistency of a summary is essential for achieving higher coherence scores. A summary that introduces multiple unrelated themes or topics, resulting in confusion and lack of focus, would likely receive a score of one. \dots \\
\midrule
\midrule
\textbf{Example Hypotheses (Decomposed Dimensions) on HANNA - EG} \\
\midrule
- The originality and creativity of the story's premise and execution are crucial for engagement. A score of 1 is given to stories that are entirely derivative, relying on clichés and predictable plots \dots \\
- The clarity and coherence of the narrative structure will significantly affect engagement scores. A score of 1 will be assigned to stories that are chaotic and incoherent \dots \\
- Stories that are overly simplistic and fail to follow the prompt effectively will receive a score of 1, while those that showcase original ideas and a compelling narrative voice will receive a score of 5. \\
\bottomrule
\end{tabular}
}
\caption{Example hypotheses for the coherence (CH) aspect of SummEval and the engagement (EG) aspect of HANNA, generated by \gpt. Each hypothesis is formulated as an evaluation rubric on a specific decomposed dimension for the aspect.}
\label{tab:example_hypotheses}
\end{table*}

\paragraph{Ablation Studies} To evaluate the effectiveness of both the hypothesis generation stage and the hypothesis-guided evaluation stage, we conduct two ablation studies. We first study the performance of \hypoeval when hypothesis generation from human evaluations and relevant literature is replaced by hypothesis generation purely from an LLM's prior knowledge. Specifically, we consider 0-shot hypothesis generation, where we directly prompt LLM $\mathcal{M}$ to generate $H_{\text{eval}}$ hypotheses for evaluating specific aspects of a text generation task, and then perform hypothesis-guided text evaluation.

We also study the effectiveness of the hypothesis-guided evaluation stage that first uses different hypotheses to generate scores and then combines them with a checklist-like approach. Specifically, we test against a pipeline similar to~\citet{liu2023calibratingllmbasedevaluator}, where we concatenated all $H_{\text{eval}}$ hypotheses after the hypothesis selection stage into one single criterion and let the evaluator model directly assign scores on a given text based on the criterion.

As shown in \cref{tab:combined_ablation}, we observe performance drops in most settings when either the hypothesis generation stage or the hypothesis-guided evaluation stage is removed. On average across both models, all settings, and both correlations, replacing the hypothesis generation stage by 0-shot generation drops performance by 7.25\%, replacing the hypothesis-guided evaluation stage with single criterion drops the performance by 8.19\%. 

\begin{table*}[t]
    \centering
    \centering
\resizebox{1\textwidth}{!}{%
\begin{tabular}{@{}llcccccccccccccccc@{}}

\toprule

 &  & \multicolumn{8}{c}{SummEval} & \multicolumn{8}{c}{NewsRoom} \\

\cmidrule(lr){3-10} \cmidrule(lr){11-18}

Models &  Methods & \multicolumn{2}{c}{CH} & \multicolumn{2}{c}{CON} & \multicolumn{2}{c}{FLU} & \multicolumn{2}{c}{RE} & \multicolumn{2}{c}{CH} & \multicolumn{2}{c}{INF} & \multicolumn{2}{c}{FLU} & \multicolumn{2}{c}{RE}\\

\cmidrule(lr){3-4} \cmidrule(lr){5-6} \cmidrule(lr){7-8} \cmidrule(lr){9-10} \cmidrule(lr){11-12} \cmidrule(lr){13-14} \cmidrule(lr){15-16} \cmidrule(lr){17-18}

& & $\rho$ & $r$ & $\rho$ & $r$ & $\rho$ & $r$ & $\rho$ & $r$ & $\rho$ & $r$ & $\rho$ & $r$ & $\rho$ & $r$ & $\rho$ & $r$ \\
\midrule
\multirow{3}{*}{\gptsmall} \\
    & \hypoeval & 0.58 & 0.58 & 0.51 & 0.62 & 0.40 & 0.45 & 0.54 & 0.58 & 0.64 & 0.69 & 0.62 & 0.75 & 0.67 & 0.69 & 0.60 & 0.78 \\
    & 0-shot generation & 0.55 \decrease & 0.56 \decrease & 0.49 \decrease & 0.58 \decrease & 0.37 \decrease & 0.40 \decrease & 0.54 \nochange & 0.56 \decrease & 0.59 \decrease & 0.64 \decrease & 0.63 \increase & 0.71 \decrease & 0.58 \decrease & 0.61 \decrease & 0.57 \decrease & 0.72 \decrease \\
    & Single criterion & 0.52 \decrease & 0.52 \decrease & 0.50 \decrease & 0.54 \decrease & 0.32 \decrease & 0.35 \decrease & 0.47 \decrease & 0.50 \decrease & 0.61 \decrease & 0.63 \decrease & 0.64 \increase & 0.70 \decrease & 0.60 \decrease & 0.59 \decrease & 0.62 \increase & 0.72 \decrease \\

\midrule
\multirow{3}{*}{\llamasmall} \\
    & \hypoeval & 0.63 & 0.63 & 0.49 & 0.62 & 0.35 & 0.35 & 0.54 & 0.56 & 0.62 & 0.65 & 0.65 & 0.74 & 0.65 & 0.64 & 0.52 & 0.73 \\
    & 0-shot generation & 0.62 \decrease & 0.64 \increase & 0.51 \increase & 0.67 \increase & 0.24 \decrease & 0.24 \decrease & 0.47 \decrease & 0.50 \decrease & 0.60 \decrease & 0.65 \nochange & 0.63 \decrease & 0.75 \increase & 0.51 \decrease & 0.50 \decrease & 0.50 \decrease & 0.73 \nochange \\
    & Single criterion & 0.50 \decrease & 0.51 \decrease & 0.48 \decrease & 0.53 \decrease & 0.37 \increase & 0.41 \increase & 0.46 \decrease & 0.48 \decrease & 0.57 \decrease & 0.58 \increase & 0.60 \decrease & 0.67 \decrease & 0.61 \decrease & 0.61 \decrease & 0.53 \increase & 0.65 \decrease \\
\bottomrule
\end{tabular}
}
\label{tab:ablation_01}

    \bigskip
    \centering
\resizebox{1\textwidth}{!}{%
\begin{tabular}{@{}llcccccccccccccccccccc@{}}

\toprule

 & & \multicolumn{12}{c}{HANNA} & \multicolumn{8}{c}{WritingPrompt-A} \\

\cmidrule(lr){3-14} \cmidrule(lr){15-22}

Models & Methods  & \multicolumn{2}{c}{CH} & \multicolumn{2}{c}{CX} & \multicolumn{2}{c}{EM} & \multicolumn{2}{c}{EG} & \multicolumn{2}{c}{RE} & \multicolumn{2}{c}{SU} & \multicolumn{2}{c}{GRA} & \multicolumn{2}{c}{COH} & \multicolumn{2}{c}{LIK} & \multicolumn{2}{c}{RE}\\

\cmidrule(lr){3-4} \cmidrule(lr){5-6} \cmidrule(lr){7-8} \cmidrule(lr){9-10} \cmidrule(lr){11-12} \cmidrule(lr){13-14} \cmidrule(lr){15-16} \cmidrule(lr){17-18} \cmidrule(lr){19-20} \cmidrule(lr){21-22}

& & $\rho$ & $r$ & $\rho$ & $r$ & $\rho$ & $r$ & $\rho$ & $r$ & $\rho$ & $r$ & $\rho$ & $r$ & $\rho$ & $r$ & $\rho$ & $r$ & $\rho$ & $r$ & $\rho$ & $r$\\

\midrule
\multirow{2}{*}{\gptsmall} \\
    & \hypoeval & 0.55 & 0.68 & 0.55 & 0.61 & 0.50 & 0.57 & 0.54 & 0.63 & 0.49 & 0.62 & 0.38 & 0.45 & 0.54 & 0.53 & 0.64 & 0.60 & 0.53 & 0.52 & 0.70 & 0.68 \\
    & 0-shot generation & 0.46 \decrease & 0.55 \decrease & 0.46 \decrease & 0.41 \decrease & 0.47 \decrease & 0.40 \decrease & 0.51 \decrease & 0.50 \decrease & 0.58 \increase & 0.57 \decrease & 0.41 \increase & 0.27 \decrease & 0.54 \nochange & 0.54 \increase & 0.62 \decrease & 0.57 \decrease & 0.45 \decrease & 0.43 \decrease & 0.71 \increase & 0.70 \increase \\
    & Single criterion & 0.49 \decrease & 0.59 \decrease & 0.56 \increase & 0.63 \increase & 0.44 \decrease & 0.49 \decrease & 0.50 \decrease & 0.57 \decrease & 0.48 \decrease & 0.57 \decrease & 0.42 \increase & 0.46 \increase & 0.44 \decrease & 0.42 \decrease & 0.57 \decrease & 0.54 \decrease & 0.50 \decrease & 0.48 \decrease & 0.65 \decrease & 0.65 \decrease \\

\midrule
\multirow{2}{*}{\llamasmall} \\
    & \hypoeval & 0.54 & 0.67 & 0.56 & 0.66 & 0.47 & 0.54 & 0.52 & 0.60 & 0.51 & 0.63 & 0.40 & 0.50 & 0.44 & 0.41 & 0.63 & 0.62 & 0.53 & 0.51 & 0.69 & 0.68 \\
    & 0-shot generation & 0.52 \decrease & 0.65 \decrease & 0.50 \decrease & 0.56 \decrease & 0.45 \decrease & 0.51 \decrease & 0.50 \decrease & 0.57 \decrease & 0.58 \increase & 0.68 \increase & 0.35 \decrease & 0.38 \decrease & 0.46 \increase & 0.43 \increase & 0.61 \decrease & 0.59 \decrease & 0.37 \decrease & 0.36 \decrease & 0.70 \increase & 0.70 \increase \\
    & Single criterion & 0.49 \decrease & 0.60 \decrease & 0.54 \decrease & 0.63 \decrease & 0.43 \decrease & 0.51 \decrease & 0.47 \decrease & 0.55 \decrease & 0.50 \decrease & 0.60 \decrease & 0.42 \increase & 0.51 \increase & 0.39 \decrease & 0.35 \decrease & 0.59 \decrease & 0.58 \decrease & 0.51 \decrease & 0.50 \decrease & 0.66 \decrease & 0.66 \decrease \\

\bottomrule
\end{tabular}
}
\label{tab:ablation_02}

    \caption{Evaluation results of the ablation studies. We use \increase and \decrease to indicate performance changes relative to \hypoeval, where \increase denotes an increase and \decrease a decrease, or \nochange for no significant change.}
    \label{tab:combined_ablation}
\end{table*}

\paragraph{Out-of-distribution generalizability.} In this section, we will demonstrate the out-of-distribution (OOD) generalizability of \hypoeval on different datasets. We conducted a cross-dataset study that uses hypotheses generated from one dataset on another (OOD) dataset of the same task. 

Specifically, for the summarization task, we use hypotheses generated from SummEval to perform hypothesis-guided evaluation for NewsRoom and vice versa on 3 aspects: coherence, fluency, and relevance. For the story generation task, we use hypotheses generated from HANNA to perform evaluation for WritingPrompt-A and vice versa on coherence or cohesiveness, and relevance. As shown in \cref{tab:generalizability}, the hypotheses generated from one dataset can be effectively used for hypothesis-guided evaluation on an OOD dataset of the same task, with an average performance change of less than 1\% for both models.

\begin{table*}[t]
\centering
\resizebox{1\textwidth}{!}{%
\begin{tabular}{@{}llcccccccccccccccccccc@{}}

\toprule

 &  & \multicolumn{6}{c}{SummEval} & \multicolumn{6}{c}{NewsRoom} & \multicolumn{4}{c}{HANNA} & \multicolumn{4}{c}{WritingPrompt-A} \\

\cmidrule(lr){3-8} \cmidrule(lr){9-14} \cmidrule(lr){15-18} \cmidrule(lr){19-22}

Models & Methods & \multicolumn{2}{c}{CH} & \multicolumn{2}{c}{FLU} & \multicolumn{2}{c}{RE} & \multicolumn{2}{c}{CH} & \multicolumn{2}{c}{FLU} & \multicolumn{2}{c}{RE} & \multicolumn{2}{c}{CH/COH} & \multicolumn{2}{c}{RE} & \multicolumn{2}{c}{CH/COH} & \multicolumn{2}{c}{RE} \\

\cmidrule(lr){3-4} \cmidrule(lr){5-6} \cmidrule(lr){7-8} \cmidrule(lr){9-10} \cmidrule(lr){11-12} \cmidrule(lr){13-14} \cmidrule(lr){15-16} \cmidrule(lr){17-18} \cmidrule(lr){19-20} \cmidrule(lr){21-22}

& & $\rho$ & $r$ & $\rho$ & $r$ & $\rho$ & $r$ & $\rho$ & $r$ & $\rho$ & $r$ & $\rho$ & $r$ & $\rho$ & $r$ & $\rho$ & $r$ & $\rho$ & $r$ & $\rho$ & $r$\\

\midrule
\multirow{2}{*}{\gptsmall}
    & IND \hypoeval & 0.58 & 0.58 & 0.40 & 0.45 & 0.54 & 0.58 & 0.64 & 0.69 & 0.67 & 0.69 & 0.60 & 0.78 & 0.55 & 0.68 & 0.49 & 0.62 & 0.64 & 0.60 & 0.70 & 0.68 \\
    & OOD \hypoeval & 0.59 & 0.61 & 0.42 & 0.44 & 0.53 & 0.56 & 0.63 & 0.68 & 0.70 & 0.73 & 0.60 & 0.77 & 0.56 & 0.68 & 0.54 & 0.65 & 0.64 & 0.59 & 0.68 & 0.66 \\

\midrule
\multirow{2}{*}{\llamasmall}
    & IND \hypoeval & 0.63 & 0.63 & 0.35 & 0.35 & 0.54 & 0.56 & 0.62 & 0.65 & 0.65 & 0.64 & 0.52 & 0.73 & 0.54 & 0.67 & 0.51 & 0.63 & 0.63 & 0.62 & 0.69 & 0.68 \\
    & OOD \hypoeval & 0.62 & 0.63 & 0.31 & 0.36 & 0.51 & 0.53 & 0.62 & 0.65 & 0.67 & 0.66 & 0.53 & 0.69 & 0.53 & 0.66 & 0.54 & 0.65 & 0.61 & 0.60 & 0.70 & 0.70 \\
\bottomrule
\end{tabular}
}
\caption{Results for OOD generalizability study, where columns are the settings that \hypoeval evaluates on. IND \hypoeval refers to hypothesis generation using in-distribution (IND) training data, while OOD \hypoeval refers to using OOD data.} 
\label{tab:generalizability}
\end{table*}

\paragraph{Cross-model} To further assess the generalizability of the decomposed dimensions, we conduct a cross-model study, where hypotheses generated by one model $\mathcal{M}_G$ are used for evaluation by another model $\mathcal{M}_E$. Results are shown in \cref{tab:combined_cross_model}. On average, for hypotheses generated by \gpt, changing the evaluator model to \llama leads to a 2.0\% drop in performance; for \llama as the generator model, changing the evaluator model to \gpt increases the performance by 1.37\%. This shows that hypotheses can be effectively transferred to different evaluator models.

\begin{table*}[t]
    \centering
    \centering
\resizebox{1\textwidth}{!}{%
\begin{tabular}{@{}llcccccccccccccccc@{}}

\toprule

 &  & \multicolumn{8}{c}{SummEval} & \multicolumn{8}{c}{NewsRoom} \\

\cmidrule(lr){3-10} \cmidrule(lr){11-18}

$\mathcal{M}_G$ &  $\mathcal{M}_E$ & \multicolumn{2}{c}{CH} & \multicolumn{2}{c}{CON} & \multicolumn{2}{c}{FLU} & \multicolumn{2}{c}{RE} & \multicolumn{2}{c}{CH} & \multicolumn{2}{c}{INF} & \multicolumn{2}{c}{FLU} & \multicolumn{2}{c}{RE}\\

\cmidrule(lr){3-4} \cmidrule(lr){5-6} \cmidrule(lr){7-8} \cmidrule(lr){9-10} \cmidrule(lr){11-12} \cmidrule(lr){13-14} \cmidrule(lr){15-16} \cmidrule(lr){17-18}

& & $\rho$ & $r$ & $\rho$ & $r$ & $\rho$ & $r$ & $\rho$ & $r$ & $\rho$ & $r$ & $\rho$ & $r$ & $\rho$ & $r$ & $\rho$ & $r$ \\
\midrule
\multirow{2}{*}{\gptsmall} \\
    & \gptsmall & 0.58 & 0.58 & 0.51 & 0.63 & 0.40 & 0.45 & 0.54 & 0.58 & 0.64 & 0.69 & 0.62 & 0.75 & 0.67 & 0.69 & 0.60 & 0.78 \\
    & \llamasmall & 0.63 & 0.66 & 0.50 & 0.60 & 0.38 & 0.44 & 0.56 & 0.59 & 0.59 & 0.66 & 0.59 & 0.75 & 0.68 & 0.69 & 0.54 & 0.74 \\

\midrule
\multirow{2}{*}{\llamasmall} \\
    & \llamasmall & 0.63 & 0.63 & 0.49 & 0.62 & 0.35 & 0.35 & 0.54 & 0.56 & 0.62 & 0.65 & 0.65 & 0.74 & 0.65 & 0.64 & 0.52 & 0.73 \\
    & \gptsmall & 0.56 & 0.55 & 0.50 & 0.61 & 0.40 & 0.41 & 0.52 & 0.56 & 0.65 & 0.70 & 0.60 & 0.71 & 0.66 & 0.68 & 0.58 & 0.75 \\
\bottomrule
\end{tabular}
}
\label{tab:cross_model_01}

    \bigskip
    \centering
\resizebox{1\textwidth}{!}{%
\begin{tabular}{@{}llcccccccccccccccccccc@{}}

\toprule

 & & \multicolumn{12}{c}{HANNA} & \multicolumn{8}{c}{WritingPrompt-A} \\

\cmidrule(lr){3-14} \cmidrule(lr){15-22}

$\mathcal{M}_G$ &  $\mathcal{M}_E$  & \multicolumn{2}{c}{CH} & \multicolumn{2}{c}{CX} & \multicolumn{2}{c}{EM} & \multicolumn{2}{c}{EG} & \multicolumn{2}{c}{RE} & \multicolumn{2}{c}{SU} & \multicolumn{2}{c}{GRA} & \multicolumn{2}{c}{COH} & \multicolumn{2}{c}{LIK} & \multicolumn{2}{c}{RE}\\

\cmidrule(lr){3-4} \cmidrule(lr){5-6} \cmidrule(lr){7-8} \cmidrule(lr){9-10} \cmidrule(lr){11-12} \cmidrule(lr){13-14} \cmidrule(lr){15-16} \cmidrule(lr){17-18} \cmidrule(lr){19-20} \cmidrule(lr){21-22}

& & $\rho$ & $r$ & $\rho$ & $r$ & $\rho$ & $r$ & $\rho$ & $r$ & $\rho$ & $r$ & $\rho$ & $r$ & $\rho$ & $r$ & $\rho$ & $r$ & $\rho$ & $r$ & $\rho$ & $r$\\

\midrule
\multirow{2}{*}{\gptsmall} \\
    & \gptsmall & 0.55 & 0.68 & 0.55 & 0.61 & 0.50 & 0.57 & 0.54 & 0.63 & 0.49 & 0.62 & 0.38 & 0.45 & 0.54 & 0.53 & 0.64 & 0.60 & 0.53 & 0.52 & 0.70 & 0.68 \\
    & \llamasmall & 0.48 & 0.65 & 0.54 & 0.65 & 0.50 & 0.57 & 0.53 & 0.61 & 0.53 & 0.65 & 0.41 & 0.49 & 0.42 & 0.39 & 0.60 & 0.58 & 0.56 & 0.54 & 0.65 & 0.65 \\

\midrule
\multirow{2}{*}{\llamasmall} \\
    & \llamasmall & 0.54 & 0.67 & 0.56 & 0.66 & 0.47 & 0.54 & 0.52 & 0.60 & 0.51 & 0.63 & 0.40 & 0.50 & 0.44 & 0.41 & 0.63 & 0.62 & 0.53 & 0.51 & 0.69 & 0.69 \\
    & \gptsmall & 0.56 & 0.69 & 0.54 & 0.61 & 0.49 & 0.57 & 0.54 & 0.62 & 0.49 & 0.60 & 0.40 & 0.49 & 0.51 & 0.50 & 0.62 & 0.59 & 0.54 & 0.53 & 0.71 & 0.71 \\
\bottomrule
\end{tabular}
}
\label{tab:cross_model_02}

    \caption{Results for cross-model study, where the hypotheses generated by one model are used for evaluation with different evaluator models.}
    \label{tab:combined_cross_model}
\end{table*}

\begin{figure*}[t]
    \centering
    \includegraphics[width=1\textwidth]{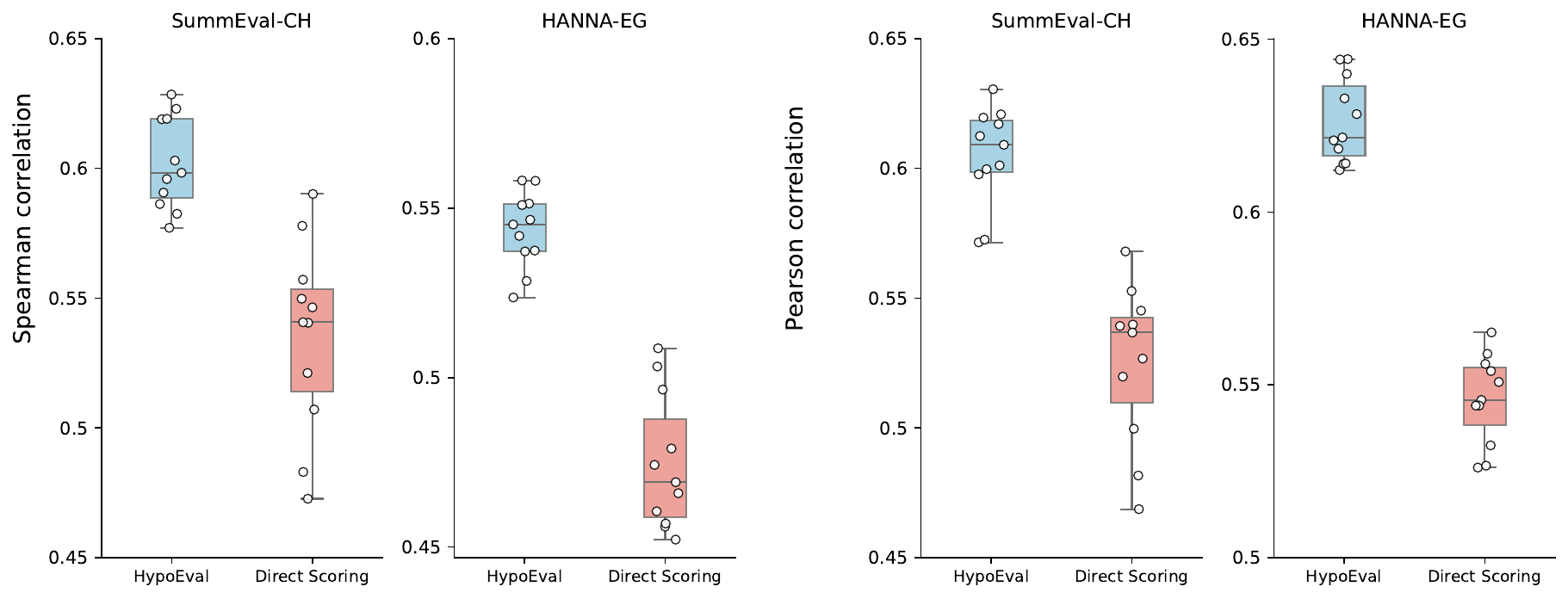}
    \caption{Results of prompt robustness study comparing \hypoeval with direct scoring, where each dot in the box plots refers to a specific prompt variation. \hypoeval shows significantly stronger robustness to evaluation prompts on representative evaluation settings.}
    \label{fig:prompt_robustness}
\end{figure*}

\paragraph{Prompt robustness.} As LLM-as-a-judge methods often exhibit high prompt sensitivity \citep{zhou2024fairerpreferenceselicitimproved, sclar2024quantifyinglanguagemodelssensitivity}, we analyze the robustness of \hypoeval to variations in evaluation instructions and compare it with direct scoring with automatic 
chain-of-thought prompting. We use GPT-4o \citep{openai2023gpt4} to generate 10 variations of the initial evaluation prompt for both \hypoeval and direct scoring that are used in the main experiments. We then perform evaluation with \gpt on the coherence (CH) aspect for SummEval and the engagement (EG) aspect for HANNA to showcase prompt robustness for the two tasks (\cref{fig:prompt_robustness}). \hypoeval shows significantly lower sensitivity to evaluation prompt variations on both settings and both meta-evaluation metrics, on average reducing the spread of Spearman correlation and Pearson correlation by 47.5\% and 29.2\%.

This study demonstrates that \hypoeval is robust to prompt variations. This robustness of \hypoeval may stem from its decomposed dimensions, which often reduce subjectivity in automated evaluation and are therefore less sensitive to variations in instructions that reinterpret the evaluated aspect.

\section{Related Work}
\label{sec:related_work}
\para{LLMs as evaluators.}
Our work follows the extensive research line on utilizing language models to automatically evaluate natural language generations  \citep{liu2023gevalnlgevaluationusing, fu2023gptscoreevaluatedesire, chen2023exploringuselargelanguage, chiang2023largelanguagemodelsalternative, li2025generationjudgmentopportunitieschallenges}. LLM evaluators are usually cheaper than human evaluations and have better alignment with human judgments than lexical metrics. GPTScore \citep{fu2023gptscoreevaluatedesire} utilizes generated pre-trained models and formulates automatic evaluation as a conditional generation task. G-Eval \citep{liu2023gevalnlgevaluationusing} similarly uses pre-trained models but adopts a prompt-based scoring approach. \citet{liu-etal-2024-calibrating, li2024calibraevalcalibratingpredictiondistribution, li2024splitmergealigningposition} consider calibration methods to mitigate the inference bias when using LLMs to assign scores. Specifically, \citet{liu-etal-2024-calibrating} utilizes an approach similar to ours by prompting LLMs to generate scoring criteria from Monte-Carlo samples. However, their framework requires a much larger corpus of ground-truth samples (e.g. up to 188 samples for summarization) and there is no publicly available code.  Alternatively, a significant amount of research has focused on automatic evaluation as a pairwise ranking problem \citep{qin2024largelanguagemodelseffective, liusie2024llmcomparativeassessmentzeroshot}. \citet{liu2025aligninghumanjudgementrole} develops an uncertainty-guided search method for ranking text generations, but is limited to an offline evaluation setting. \citet{zhou2024fairerpreferenceselicitimproved} introduces a prompt optimization framework that elicits both fairer preferences and better alignment with humans. However, pairwise ranking evaluation can face problems in terms of scalability, online evaluation, and cost or efficiency issues.

\para{Checklist-based evaluation.}
Similar to our hypothesis-guided evaluation, where we aggregate scores from each decomposed dimension to acquire an overall score, there has also been previous research on aggregating evaluation results on atomic checklists for better correlation with human judgments. However, the checklist line of work mainly focuses on binary checklists where the answer is restricted to YES or NO. \citet{tan2024proxyqaalternativeframeworkevaluating, que2024hellobenchevaluatinglongtext, zhou2023instructionfollowingevaluationlargelanguage, qin2024infobenchevaluatinginstructionfollowing} use human-curated proxy questions, checklists, or "verifiable instructions" to benchmark LLMs' long-form text generation or instruction-following capabilities. \citet{cook2024tickingboxesgeneratedchecklists} explores automatic checklist generation by prompting with few-shot templates and shows effectiveness in evaluating instruction-following. \citet{lee2024checkevalrobustevaluationframework} and \citet{pereira2024checkevalchecklistbasedapproachevaluating} further utilize the binary checklist method on evaluating natural language text generations such as summarization, but on average it does not yield better results than non-checklist methods like G-Eval.

\section{Conclusion}
\label{sec:conclusion}

We propose \hypoeval, a tuning-free and sample-efficient automated evaluation framework for natural language generation that achieves state-of-the-art performance in alignment with human evaluation rankings and scores. The generated hypotheses serve as decomposed dimensions of desiderate and provide interpretable explanations of the automated evaluation process. Through systematic studies, we show the robustness of \hypoeval to OOD data, prompt variations, and different evaluator models.

\clearpage
\appendix

\section{Prompts}
\label{appendix:prompts}

We include some example prompts for both the hypothesis generation and the hypothesis-guided evaluation stages of \hypoeval.

\subsection{Summarization}
\label{appendix:prompts:summarization}

\begin{lstlisting}[caption={Hypothesis Generation.},label={lst:summ:hypgen},firstnumber=auto]
@\textcolor{mygray}{Instruction Prompt}@
You are a helpful assistant for predicting what score on @\textcolor{mypurple}{<aspect>}@, between 1 to 5 (the higher the better), will a summary of a passage receive when judged by human experts.
Given a set of summaries and their source texts, we want to generate hypotheses that are useful for predicting what score on @\textcolor{mypurple}{<aspect>}@, between 1 to 5 (the higher the better), will a summary of a passage receive when judged by human experts.
The definition of @\textcolor{mypurple}{<aspect>}@ is given by: @\textcolor{mypurple}{<definition>}@

Using the given examples and relevant literatures, please propose @\textcolor{mypurple}{<num\_hypotheses>}@ possible hypotheses.
These hypotheses should identify specific patterns that occur across the provided summaries.

Each hypothesis should be about a specific trait or dimension that human experts considers when giving score on @\textcolor{mypurple}{<aspect>}@.
Each hypothesis should clearly state that based on the trait or dimension, what kind of summary would be given a score of one, what kind of summary a score of two, what kind of summary a score of three, what kind of summary a score of four, and what kind of summary a score of five.

Generate them in the format of hypothesis1. [hypothesis], hypothesis2. [hypothesis], ... hypothesis @\textcolor{mypurple}{<num\_hypotheses>}@. [hypothesis].
The hypotheses should analyze what are the traits of the summaries human experts considers when giving a score of one, two, three, four, or five.
Remember! when generating hypotheses, always put "hypothesis1.", "hypothesis2.", etc. as your index, do not just generate "1.", "2.", etc.

@\textcolor{mygray}{User Prompt}@
We have seen some summaries and their source texts, together with their scores on @\textcolor{mypurple}{<aspect>}@ given by human experts:
@\textcolor{mypurple}{<observations>}@
Please generate hypotheses that are useful for predicting what score on @\textcolor{mypurple}{<aspect>}@, between 1 to 5 (the higher the better), will a summary of a passage receive when judged by human experts.
The definition of @\textcolor{mypurple}{<aspect>}@ is given by: @\textcolor{mypurple}{<definition>}@
Propose @\textcolor{mypurple}{<num\_hypotheses>}@ possible hypotheses. Generate them in the format of hypothesis1. [hypothesis], hypothesis2. [hypothesis], ... hypothesis @\textcolor{mypurple}{<num\_hypotheses>}@. [hypothesis].
Remember! when generating hypotheses, always put "hypothesis1.", "hypothesis2.", etc. as your index, do not just generate "1.", "2.", etc.
Proposed hypotheses:
\end{lstlisting}

\begin{lstlisting}[caption={Hypothesis Refine with Data.},label={lst:summ:refine_data},firstnumber=auto]
@\textcolor{mygray}{Instruction Prompt}@
You are a helpful assistant for predicting what score on @\textcolor{mypurple}{<aspect>}@, between 1 to 5 (the higher the better), will a summary of a passage receive when judged by human experts.
Given a set of summaries and their source texts, we want to generate hypotheses that are useful for predicting what score on @\textcolor{mypurple}{<aspect>}@, between 1 to 5 (the higher the better), will a summary of a passage receive when judged by human experts.
The definition of @\textcolor{mypurple}{<aspect>}@ is given by: @\textcolor{mypurple}{<definition>}@

Using the given examples, refine the hypotheses provided.
The desired hypotheses should identify specific patterns that occur across the provided summaries.

Each hypothesis should be about a specific trait or dimension that human experts considers when giving score on @\textcolor{mypurple}{<aspect>}@.
Each hypothesis should clearly state that based on the trait or dimension, what kind of summary would be given a score of one, what kind of summary a score of two, what kind of summary a score of three, what kind of summary a score of four, and what kind of summary a score of five.

Generate them in the format of hypothesis1. [hypothesis], hypothesis2. [hypothesis], ... @\textcolor{mypurple}{<num\_hypotheses>}@. [hypothesis].
The hypotheses should analyze what are the traits of the summaries human experts considers when giving a score of one, two, three, four, or five.
Remember! when generating hypotheses, always put "hypothesis1.", "hypothesis2.", etc. as your index, do not just generate "1.", "2.", etc.
@\textcolor{mygray}{User Prompt}@
We have seen some summaries and their source texts, together with their scores on @\textcolor{mypurple}{<aspect>}@ given by human experts:
@\textcolor{mypurple}{<observations>}@
We have some hypotheses need to be refined:
@\textcolor{mypurple}{<hypotheses>}@
Please refine these hypotheses to make them more specific and useful for predicting what score on @\textcolor{mypurple}{<aspect>}@, between 1 to 5 (the higher the better), will a summary of a passage receive when judged by human experts.
When refining the hypotheses, feel free to change the key information or topic of a hypothesis based on the provided prevailing patterns in data if you think it is necessary.
Generate the refined hypotheses in the format of hypothesis1. [hypothesis], hypothesis2. [hypothesis], ... hypothesis @\textcolor{mypurple}{<num\_hypotheses>}@. [hypothesis].
The refined hypotheses should analyze what are the traits of the summaries human experts considers when giving a score of one, two, three, four, or five.
Remember! when generating the refined hypotheses, always put "hypothesis1.", "hypothesis2.", etc. as your index, do not just generate "1.", "2.", etc.
Refined hypotheses:
\end{lstlisting}

\begin{lstlisting}[caption={Hypothesis Refine with Literature.},label={lst:summ:refine_paper},firstnumber=auto]
@\textcolor{mygray}{Instruction Prompt}@
You are a helpful assistant for predicting what score on @\textcolor{mypurple}{<aspect>}@, between 1 to 5 (the higher the better), will a summary of a passage receive when judged by human experts.
Given a set of summaries and their source texts, we want to generate hypotheses that are useful for predicting what score on @\textcolor{mypurple}{<aspect>}@, between 1 to 5 (the higher the better), will a summary of a passage receive when judged by human experts.
The definition of @\textcolor{mypurple}{<aspect>}@ is given by: @\textcolor{mypurple}{<definition>}@

Using the given relevant literatures, refine the hypotheses provided.
The desired hypotheses should identify specific patterns that occur across the provided summaries.

Each hypothesis should be about a specific trait or dimension that human experts considers when giving score on @\textcolor{mypurple}{<aspect>}@.
Each hypothesis should clearly state that based on the trait or dimension, what kind of summary would be given a score of one, what kind of summary a score of two, what kind of summary a score of three, what kind of summary a score of four, and what kind of summary a score of five.

Generate them in the format of hypothesis1. [hypothesis], hypothesis2. [hypothesis], ... hypothesis @\textcolor{mypurple}{<num\_hypotheses>}@. [hypothesis].
The hypotheses should analyze what are the traits of the summaries human experts considers when giving a score of one, two, three, four, or five.
Remember! when generating hypotheses, always put "hypothesis1.", "hypothesis2.", etc. as your index, do not just generate "1.", "2.", etc.
@\textcolor{mygray}{User Prompt}@
We have some key findings from a series of research papers that might be useful for generating hypotheses:
@\textcolor{mypurple}{<relevant\_papers>}@
We have some hypotheses need to be refined:
@\textcolor{mypurple}{<hypotheses>}@
Please refine these hypotheses to make them more specific and useful for predicting what score on @\textcolor{mypurple}{<aspect>}@, between 1 to 5 (the higher the better), will a summary of a passage receive when judged by human experts.
When refining the hypotheses, feel free to change the key information or topic of a hypothesis based on the provided prevailing patterns in data if you think it is necessary.
Generate the refined hypotheses in the format of hypothesis1. [hypothesis], hypothesis2. [hypothesis], ... hypothesis @\textcolor{mypurple}{<num\_hypotheses>}@. [hypothesis].
The refined hypotheses should analyze what are the traits of the summaries human experts considers when giving a score of one, two, three, four, or five.
Remember! when generating the refined hypotheses, always put "hypothesis1.", "hypothesis2.", etc. as your index, do not just generate "1.", "2.", etc.
Refined hypotheses:
\end{lstlisting}

\begin{lstlisting}[caption={Check Hypothesis Repetition},label={lst:summ:check_repetition},firstnumber=auto]
@\textcolor{mygray}{Instruction Prompt}@
You are a helpful assistant for predicting what score on @\textcolor{mypurple}{<aspect>}@, between 1 to 5 (the higher the better), will a summary of a passage receive when judged by human experts.

From past experiences, you learned two hypotheses that are useful for predicting what score on @\textcolor{mypurple}{<aspect>}@, between 1 to 5 (the higher the better), will a summary of a passage receive when judged by human experts.

You need to determine if the two hypotheses are so similar to the level of "repeating hypotheses".
Finally, answer "yes" if the two hypotheses are repetitive and "no" if they are not.
Keep your answer short.

Give your final answer in the format of "Final answer: [answer]".

@\textcolor{mygray}{User Prompt}@
We have two hypotheses that need you to determine if they are repetitive:

@\textcolor{mypurple}{<hypotheses>}@
Are these two hypotheses so similar to the level that they are repetitive? If the both of them can provide significantly more information than only one of them could, and the information is important and useful for predicting what score on @\textcolor{mypurple}{<aspect>}@, between 1 to 5 (the higher the better), will a summary of a passage receive when judged by human experts, they should not be considered repetitive.

Note that adding specific examples does not count as "provide significantly more information".

Give a short explanation of your decision.
Then give your final answer in the format of "Final answer: [answer]".
Your answer: 
\end{lstlisting}

\begin{lstlisting}[caption={Hypothesis-Guided Evaluation},label={lst:summ:hyp_eval},firstnumber=auto]
@\textcolor{mygray}{Instruction Prompt}@
You are a helpful assistant in answering questions about a summary of a story.
You will be given the story, the summary, and a pattern that talks about a specific trait to evaluate the @\textcolor{mypurple}{<aspect>}@ of the summary.
You should be generous and not too strict when evaluating.
The definition of @\textcolor{mypurple}{<aspect>}@ is given by: @\textcolor{mypurple}{<definition>}@.
Story: [story]
Summary: [summary]
Pattern: [hypothesis]
The pattern talks about a specific trait that is related to the summary's score on @\textcolor{mypurple}{<aspect>}@.
You need to evaluate the summary based on the trait and the rubric that the pattern talks about. 
You should give a score (ranging from 1 to 5) on that trait according to the rubric.
Give your final evaluation score in the format of {Final score: [your score]}.

@\textcolor{mygray}{User Prompt}@
Given story, summary, and pattern:
Story: @\textcolor{mypurple}{<story>}@
Summary: @\textcolor{mypurple}{<summary>}@
Pattern: @\textcolor{mypurple}{<hypothesis>}@
The pattern talks about a specific trait that is related to the summary's score on @\textcolor{mypurple}{<aspect>}@.
The definition of @\textcolor{mypurple}{<aspect>}@ is given by: @\textcolor{mypurple}{<definition>}@
You need to evaluate the summary based on the trait and the rubric that the pattern talks about. 
You should give a score (ranging from 1 to 5) on that trait according to the rubric.
Follow the steps and provide reasoning when giving your score.
Step 1: What is the trait that the pattern talks about?
Step 2: Based on the trait and the rubric provided in the pattern, how is the summary on the trait?
Step 3 (final answer): Based on the rubric and your evaluations in step 2, what should be the score of the summary on the trait?
You should be generous and not too strict when evaluating.
Give your final evaluation score in the format of {Final score: [your score]}.
Answer:
\end{lstlisting}

\subsection{Story Generation}
\label{appendix:prompts:story_generation}

\begin{lstlisting}[caption={Hypothesis Generation.},label={lst:story:hypgen},firstnumber=auto]
@\textcolor{mygray}{Instruction Prompt}@
You are a helpful assistant for predicting what score on @\textcolor{mypurple}{<aspect>}@, between 1 to 5 (the higher the better), will a written story of a given prompt receive when judged by human experts.
Given a set of stories and their prompts, we want to generate hypotheses that are useful for predicting what score on @\textcolor{mypurple}{<aspect>}@, between 1 to 5 (the higher the better), will a written story of a given prompt receive when judged by human experts.
The definition of @\textcolor{mypurple}{<aspect>}@ is given by: @\textcolor{mypurple}{<definition>}@

Using the given examples and relevant literatures, please propose @\textcolor{mypurple}{<num\_hypotheses>}@ possible hypotheses.
These hypotheses should identify specific patterns that occur across the provided stories.

Each hypothesis should be about a specific trait or dimension that human experts considers when giving score on @\textcolor{mypurple}{<aspect>}@.
Each hypothesis should clearly state that based on the trait or dimension, what kind of story would be given a score of one, what kind of story a score of two, what kind of story a score of three, what kind of story a score of four, and what kind of story a score of five.

Generate them in the format of hypothesis1. [hypothesis], hypothesis2. [hypothesis], ... hypothesis @\textcolor{mypurple}{<num\_hypotheses>}@. [hypothesis].
The hypotheses should analyze what are the traits of the stories human experts considers when giving a score of one, two, three, four, or five.
Remember! when generating hypotheses, always put "hypothesis1.", "hypothesis2.", etc. as your index, do not just generate "1.", "2.", etc.
@\textcolor{mygray}{User Prompt}@
We have some key findings from a series of research papers that might be useful for generating hypotheses:
@\textcolor{mypurple}{<relevant\_papers>}@
We have seen some stories and their prompts, together with their scores on @\textcolor{mypurple}{<aspect>}@ given by human experts:
@\textcolor{mypurple}{<observations>}@
Please generate hypotheses that are useful for predicting what score on @\textcolor{mypurple}{<aspect>}@, between 1 to 5 (the higher the better), will a written story of a given prompt receive when judged by human experts.
The definition of @\textcolor{mypurple}{<aspect>}@ is given by: @\textcolor{mypurple}{<definition>}@
Propose @\textcolor{mypurple}{<num\_hypotheses>}@ possible hypotheses. Generate them in the format of hypothesis1. [hypothesis], hypothesis2. [hypothesis], ... hypothesis @\textcolor{mypurple}{<num\_hypotheses>}@. [hypothesis].
Remember! when generating hypotheses, always put "hypothesis1.", "hypothesis2.", etc. as your index, do not just generate "1.", "2.", etc.
Proposed hypotheses:
\end{lstlisting}

\begin{lstlisting}[caption={Hypothesis Refine with Data.},label={lst:story:refine_data},firstnumber=auto]
@\textcolor{mygray}{Instruction Prompt}@
You are a helpful assistant for predicting what score on @\textcolor{mypurple}{<aspect>}@, between 1 to 5 (the higher the better), will a written story of a given prompt receive when judged by human experts.
Given a set of stories and their prompts, we want to generate hypotheses that are useful for predicting what score on @\textcolor{mypurple}{<aspect>}@, between 1 to 5 (the higher the better), will a written story of a given prompt receive when judged by human experts.
The definition of @\textcolor{mypurple}{<aspect>}@ is given by: @\textcolor{mypurple}{<definition>}@

Using the given examples, refine the hypotheses provided.
The desired hypotheses should identify specific patterns that occur across the provided stories.

Each hypothesis should be about a specific trait or dimension that human experts considers when giving score on @\textcolor{mypurple}{<aspect>}@.
Each hypothesis should clearly state that based on the trait or dimension, what kind of story would be given a score of one, what kind of story a score of two, what kind of story a score of three, what kind of story a score of four, and what kind of story a score of five.

Generate them in the format of hypothesis1. [hypothesis], hypothesis2. [hypothesis], ... hypothesis @\textcolor{mypurple}{<num\_hypotheses>}@. [hypothesis].
The hypotheses should analyze what are the traits of the stories human experts considers when giving a score of one, two, three, four, or five.
Remember! when generating hypotheses, always put "hypothesis1.", "hypothesis2.", etc. as your index, do not just generate "1.", "2.", etc.
@\textcolor{mygray}{User Prompt}@
We have seen some stories and their prompts, together with their scores on @\textcolor{mypurple}{<aspect>}@ given by human experts:
@\textcolor{mypurple}{<observations>}@
We have some hypotheses need to be refined:
@\textcolor{mypurple}{<hypotheses>}@
Please refine these hypotheses to make them more specific and useful for predicting what score on @\textcolor{mypurple}{<aspect>}@, between 1 to 5 (the higher the better), will a written story of a given prompt receive when judged by human experts.
When refining the hypotheses, feel free to change the key information or topic of a hypothesis based on the provided prevailing patterns in data if you think it is necessary.
Generate the refined hypotheses in the format of hypothesis1. [hypothesis], hypothesis2. [hypothesis], ... hypothesis @\textcolor{mypurple}{<num\_hypotheses>}@. [hypothesis].
The refined hypotheses should analyze what are the traits of the stories human experts considers when giving a score of one, two, three, four, or five.
Remember! when generating the refined hypotheses, always put "hypothesis1.", "hypothesis2.", etc. as your index, do not just generate "1.", "2.", etc.
Refined hypotheses:
\end{lstlisting}

\begin{lstlisting}[caption={Hypothesis Refine with Literature.},label={lst:story:refine_paper},firstnumber=auto]
@\textcolor{mygray}{Instruction Prompt}@
You are a helpful assistant for predicting what score on @\textcolor{mypurple}{<aspect>}@, between 1 to 5 (the higher the better), will a written story of a given prompt receive when judged by human experts.
Given a set of stories and their prompts, we want to generate hypotheses that are useful for predicting what score on @\textcolor{mypurple}{<aspect>}@, between 1 to 5 (the higher the better), will a written story of a given prompt receive when judged by human experts.
The definition of @\textcolor{mypurple}{<aspect>}@ is given by: @\textcolor{mypurple}{<definition>}@

Using the given relevant literatures, refine the hypotheses provided.
The desired hypotheses should identify specific patterns that occur across the provided stories.

Each hypothesis should be about a specific trait or dimension that human experts considers when giving score on @\textcolor{mypurple}{<aspect>}@.
Each hypothesis should clearly state that based on the trait or dimension, what kind of story would be given a score of one, what kind of story a score of two, what kind of story a score of three, what kind of story a score of four, and what kind of story a score of five.

Generate them in the format of hypothesis1. [hypothesis], hypothesis2. [hypothesis], ... hypothesis @\textcolor{mypurple}{<num\_hypotheses>}@. [hypothesis].
The hypotheses should analyze what are the traits of the stories human experts considers when giving a score of one, two, three, four, or five.
Remember! when generating hypotheses, always put "hypothesis1.", "hypothesis2.", etc. as your index, do not just generate "1.", "2.", etc.
@\textcolor{mygray}{User Prompt}@
We have some key findings from a series of research papers that might be useful for generating hypotheses:
  @\textcolor{mypurple}{<relevant\_papers>}@
  We have some hypotheses need to be refined:
  @\textcolor{mypurple}{<hypotheses>}@
  Please refine these hypotheses to make them more specific and useful for predicting what score on @\textcolor{mypurple}{<aspect>}@, between 1 to 5 (the higher the better), will a written story of a given prompt receive when judged by human experts.
  When refining the hypotheses, feel free to change the key information or topic of a hypothesis based on the provided prevailing patterns in data if you think it is necessary.
  Generate the refined hypotheses in the format of hypothesis1. [hypothesis], hypothesis2. [hypothesis], ... hypothesis @\textcolor{mypurple}{<num\_hypotheses>}@. [hypothesis].
  The refined hypotheses should analyze what are the traits of the stories human experts considers when giving a score of one, two, three, four, or five.
  Remember! when generating the refined hypotheses, always put "hypothesis1.", "hypothesis2.", etc. as your index, do not just generate "1.", "2.", etc.
  Refined hypotheses:
\end{lstlisting}

\begin{lstlisting}[caption={Check Hypothesis Repetition},label={lst:story:check_repetition},firstnumber=auto]
@\textcolor{mygray}{Instruction Prompt}@
You are a helpful assistant for predicting what score on @\textcolor{mypurple}{<aspect>}@, between 1 to 5 (the higher the better), will a written story of a given prompt receive when judged by human experts.
From past experiences, you learned two hypotheses that are useful for predicting what score on @\textcolor{mypurple}{<aspect>}@, between 1 to 5 (the higher the better), will a written story of a given prompt receive when judged by human experts.
You need to determine if the two hypotheses are so similar to the level of "repeating hypotheses".
Finally, answer "yes" if the two hypotheses are repetitive and "no" if they are not.
Keep your answer short.
Give your final answer in the format of "Final answer: [answer]".

@\textcolor{mygray}{User Prompt}@
We have two hypotheses that need you to determine if they are repetitive:
@\textcolor{mypurple}{<hypotheses>}@
Are these two hypotheses so similar to the level that they are repetitive? If the both of them can provide significantly more information than only one of them could, and the information is important and useful for predicting what score on @\textcolor{mypurple}{<aspect>}@, between 1 to 5 (the higher the better), will a written story of a given prompt receive when judged by human experts, they should not be considered repetitive.
Note that adding specific examples does not count as "provide significantly more information".
Give a short explanation of your decision.
Then give your final answer in the format of "Final answer: [answer]".
Your answer: 

\end{lstlisting}

\begin{lstlisting}[caption={Hypothesis-Guided Evaluation},label={lst:story:hyp_eval},firstnumber=auto]
@\textcolor{mygray}{Instruction Prompt}@
You are a helpful assistant in answering questions about a written story of a given prompt.
You will be given the prompt, the written story, and a pattern that talks about a specific trait to evaluate the @\textcolor{mypurple}{<aspect>}@ of the story.
You should be generous and not too strict when evaluating.
The definition of @\textcolor{mypurple}{<aspect>}@ is given by: @\textcolor{mypurple}{<definition>}@.
Prompt: [prompt]
Story: [story]
Pattern: [hypothesis]

The pattern talks about a specific trait that is related to the story's score on @\textcolor{mypurple}{<aspect>}@.
You need to evaluate the story based on the trait and the rubric that the pattern talks about. 
You should give a score (ranging from 1 to 5) on that trait according to the rubric.
Give your final evaluation score in the format of {Final score: [your score]}.

@\textcolor{mygray}{User Prompt}@
Given prompt, story, and pattern:
Prompt: @\textcolor{mypurple}{<prompt>}@
Story: @\textcolor{mypurple}{<story>}@
Pattern: @\textcolor{mypurple}{<hypothesis>}@
Note: the story may have been abruptly cut in the middle of a sentence. Please rate it as if they ended just before the unfinished sentence.
The pattern talks about a specific trait that is related to the story's score on @\textcolor{mypurple}{<aspect>}@.
The definition of @\textcolor{mypurple}{<aspect>}@ is given by: @\textcolor{mypurple}{<definition>}@

You need to evaluate the story based on the trait and the rubric that the pattern talks about. 
You should give a score (ranging from 1 to 5) on that trait according to the rubric.

Follow the steps and provide reasoning when giving your score.
Step 1: What is the trait that the pattern talks about?
Step 2: Based on the trait and the rubric provided in the pattern, how is the story on the trait?
Step 3 (final answer): Based on the rubric and your evaluations in step 2, what should be the score of the story on the trait?
You should be generous and not too strict when evaluating.
Give your final evaluation score in the format of {Final score: [your score]}.
Answer:
\end{lstlisting}

\section{Implementation Details}
\label{appendix:implementation_details}

\subsection{Implementation Details of \hypoeval and Experiments}
\label{appendix:implementation_details:hypgen}

To collect relevant literature information $\mathcal{L}$, we first prompt Grok 3 with DeepSearch \citep{xai2025grok3} to search for relevant academic papers on the two evaluation tasks (summarization and story generation) and retrieve 15 and 10 papers, respectively. Then, we use S2ORC-doc2json \citep{lo-wang-2020-s2orc} to convert the raw PDF files to a set of JSON files that contain the abstracts and main texts of the papers. Subsequently, the hypothesis generator model $\mathcal{M}_G$ is prompted to generate a summary for each JSON file. The summaries are then concatenated to get the relevant literature information $\mathcal{L}$ that is later used for hypothesis generation with data and literature.

Then in the hypothesis generation stage, we set the size of $S_{\text{init}}$ to 5, $|\mathcal{H}^{\text{init}}| = 5$, $k = 10$, $\theta = 0.5$,  $\alpha = 0.5$, $w_{\text{max}} = 10$, $N_{\text{refine}} = 6$, and $H_{\text{max}}=20$. For the hyperparameters $a$, $b$ of the reward, we let $a=1$, $b=\frac{1}{16}$ to ensure that the exploitation term is bounded in $[0,1]$. 

For hypothesis-guided evaluation, we let $H_{\text{eval}} = 5$. Following the implementation of PairS \citep{liu2025aligninghumanjudgementrole}, we set Spearman or Pearson correlation to 1 if the human annotation scores for all candidate responses of a source text or prompt are the same.

For all experiments and additional studies, excluding the prompt robustness study, we run all methods on all settings with 3 seeds: 42, 2, 114514.

\subsection{Implementation Details of Baselines}
\label{appendix:implementation_details:baselines}

For reference-based baselines, we implement ROUGE-L-F1, BERTScore-recall with default model choice for English language, UniEval, and the bart-score-cnn-src-hypo version of BARTScore. 

For fine-tuning \llamaeightb, for FT-A, we use the same training set $S_{\text{tr}}$ as \hypoeval; for FT-B, we further sample 170 data points for SummEval, NewsRoom, and HANNA or 70 data points for WritingPrompt-A from the remaining data points, excluding the test sets. We fine-tune the model for 20 epochs.

For direct scoring and G-Eval, following the setup of the original G-Eval paper \citep{liu2023gevalnlgevaluationusing}, we first let the evaluator model $\mathcal{M}_E$ generate chain-of-thought steps for evaluation, and then let $\mathcal{M}_E$ give evaluation scores of given texts. To acquire the probabilities for G-Eval, we directly retrieve token probabilities for \llama, and sample 20 times with temperature set to 1 for \gpt.

For direct scoring with few-shot demonstrations, we set the number of demonstrations $k=3$, and randomly sample annotated data points from $S_{\text{tr}}$.

For PairS-beam, we use the same hyperparameter setting across all settings, where we set $\text{beam\_size}=1$ and $\text{prob\_gap}=0.1$.

\section{Example Hypotheses}
\label{appendix:example_hypotheses}

We include full versions of more examples of generated hypotheses for SummEval - coherence, HANNA - engagement, and NewsRoom - relevance in \cref{tab:full_example_hypotheses_01}, \cref{tab:full_example_hypotheses_02}, and \cref{tab:full_example_hypotheses_03}.

\begin{table*}[t]
\centering
\small
\resizebox{1\textwidth}{!}{%
\begin{tabular}{@{}p{\textwidth}@{}}
\toprule

\textbf{Example Hypotheses on SummEval - Coherence} \\
\midrule
\textbullet{} The overall structure and organization of the summary play a vital role in determining coherence scores. Summaries that are logically organized, with a clear introduction, body, and conclusion, will score higher (4 or 5), while those that lack a coherent structure or appear haphazardly arranged will score lower (1 or 2). A well-structured summary that guides the reader through the main points will likely receive a score of 5, while a disorganized summary will score a 1.\\
\textbullet{} Summaries that maintain a consistent tone and style throughout will be rated higher for coherence (4 or 5), as this consistency aids in reader comprehension. In contrast, summaries that shift in tone or style abruptly, creating confusion or distraction for the reader, will be rated lower (1 or 2), reflecting a lack of coherence and engagement.\\
\textbullet{} Summaries that are exceptionally coherent, well-structured, and articulate, effectively conveying the main ideas and integrating them in a way that enhances understanding, will receive a score of five. \\
\textbullet{} Summaries that are poorly structured, lack logical flow, and fail to connect ideas will receive a score of one, as they may be disjointed and confusing, making it difficult for readers to follow the main ideas. \\
\textbullet{} The thematic consistency of a summary is essential for achieving higher coherence scores. A summary that introduces multiple unrelated themes or topics, resulting in confusion and lack of focus, would likely receive a score of one. A summary that partially maintains a central theme but includes several irrelevant details or tangents that distract from the main point may receive a score of two. A summary that presents a clear main theme but lacks depth or thorough development of supporting ideas, leading to a somewhat superficial understanding, might score a three. A summary that effectively ties together related ideas around a central theme, providing a coherent narrative with some depth and relevant context, would receive a score of four. Finally, a summary that maintains a singular, well-developed theme throughout, seamlessly integrating all points and enhancing the overall message with rich context and insights, would receive a score of five. \\

\bottomrule
\end{tabular}
}
\caption{Full version of additional example hypotheses generated by \gpt for the coherence aspect of SummEval. }
\label{tab:full_example_hypotheses_01}
\end{table*}

\begin{table*}[t]
\centering
\small
\resizebox{1\textwidth}{!}{%
\begin{tabular}{@{}p{\textwidth}@{}}
\toprule

\textbf{Example Hypotheses on HANNA - Engagement} \\
\midrule
\textbullet{} The originality and creativity of the story's premise and execution are crucial for engagement. A score of 1 is given to stories that are entirely derivative, relying on predictable plots without any unique elements. A score of 2 may indicate a story that includes a few original ideas but is largely uninspired and fails to captivate the reader. A score of 3 suggests a moderately creative premise that engages the reader but lacks depth or surprising twists. A score of 4 reflects a highly original story that captivates the audience with innovative concepts and engaging execution, while a score of 5 is reserved for stories that present unique, unexpected twists and thought-provoking insights that challenge the reader's expectations and provoke deeper reflection. \\
\textbullet{} The clarity and coherence of the narrative structure will significantly affect engagement scores. A score of 1 will be assigned to stories that are chaotic and incoherent, making them nearly impossible to follow; a score of 2 for stories that have a basic structure but are confusing or lack logical flow, resulting in a disjointed reading experience; a score of 3 for stories with a clear but simplistic structure that conveys the plot adequately but lacks depth; a score of 4 for stories that are well-structured, logically flowing, and maintain reader interest through effective transitions and a clear narrative arc; and a score of 5 for stories that exhibit a sophisticated and intricate structure that enhances the narrative, captivates the reader, and seamlessly integrates various plot elements, creating a compelling reading experience. \\
\textbullet{} Stories that are overly simplistic and fail to follow the prompt effectively will receive a score of 1, while those that showcase original ideas and a compelling narrative voice will receive a score of 5. \\
\textbullet{} Emotional resonance and the ability to evoke feelings in the reader are key factors in engagement scoring. Stories that fail to connect emotionally with the audience will likely receive a score of 1 or 2, while those that successfully elicit strong emotional reactions, such as joy, sadness, or suspense, will score higher (4 or 5) due to their impactful storytelling. \\
\textbullet{} The richness of character development is a key factor in determining engagement. A score of 1 is assigned to stories featuring flat, one-dimensional characters that fail to evoke any emotional connection or interest. A score of 2 may indicate characters that are somewhat developed but lack complexity and relatability, making it hard for readers to connect. A score of 3 suggests characters that are relatable but not fully fleshed out, leading to moderate engagement. A score of 4 reflects well-developed characters that enhance the overall engagement of the story, showcasing growth, complexity, and emotional depth. Conversely, stories with multi-dimensional, relatable characters that undergo meaningful development, face internal and external challenges, and elicit empathy from the reader will score a 5. \\

\bottomrule
\end{tabular}
}
\caption{Full version of additional example hypotheses generated by \gpt for the engagement aspect of HANNA.}
\label{tab:full_example_hypotheses_02}
\end{table*}

\begin{table*}[t]
\centering
\small
\resizebox{1\textwidth}{!}{%
\begin{tabular}{@{}p{\textwidth}@{}}
\toprule

\textbf{Example Hypotheses on NewsRoom - Relevance} \\
\midrule
\textbullet{} A summary will receive a score of 1 if it contains information that directly contradicts the source text, a score of 2 if it contains some information not present in the source text, a score of 3 if it contains a mix of information present and not present in the source text, a score of 4 if it contains most information present in the source text, but lacks nuance or depth, and a score of 5 if it only contains information present in the source text, has excellent coherence, clarity, and demonstrates a high level of depth and insight, with effective use of transitional phrases and sentences to connect ideas, and the summary is accurate and reliable. \\
\textbullet{} A summary will receive a score of 1 if it is completely unrelated to the source text, a score of 2 if it is partially related but contains significant inaccuracies, a score of 3 if it is partially related and contains some accurate information, but also some inaccuracies, a score of 4 if it is mostly related and contains mostly accurate information, and has good coherence and clarity, but misses some key points or lacks depth, and a score of 5 if it is entirely related to the source text, contains all accurate and key information, and demonstrates a high level of coherence, clarity, and depth, with clear and concise language, and effective use of rhetorical devices to engage the reader and convey complex ideas, and the summary is comprehensive and well-written. \\
\textbullet{} A summary will receive a score of 1 if it introduces a significant amount of new information not present in the source text, such as external knowledge or opinions, that alters the meaning or tone of the original text, a score of 2 if it introduces some new information but also includes some relevant details from the source text, a score of 3 if it includes a mix of relevant and irrelevant details with some inconsistencies, such as including information from other sources, a score of 4 if it includes mostly relevant details with minor errors or omissions, and a score of 5 if it only includes details that are present in the source text and are relevant to the main points, without any external information or opinions that could change the original meaning, based on the trait of relevance and presence of extraneous information, including the ability to distinguish between essential and non-essential information. \\
\textbullet{} A summary will receive a score of 1 if it fails to capture any key concepts or relationships presented in the source text, a score of 2 if it captures some key concepts but misses important relationships or nuances, a score of 3 if it captures most key concepts and relationships but with some inaccuracies or inconsistencies, a score of 4 if it accurately captures most key concepts and relationships with minor inaccuracies, and a score of 5 if it accurately and comprehensively captures all key concepts and relationships presented in the source text, including underlying themes, motivations, and implications, based on the trait of depth and quality of analysis, including the ability to identify and explain complex relationships, patterns, and concepts presented in the source text.\\

\bottomrule
\end{tabular}
}
\caption{Full version of additional example hypotheses generated by \llama for the relevance aspect of NewsRoom.}
\label{tab:full_example_hypotheses_03}
\end{table*}

\section{Additional Illustrations}
\label{appendix:additional_illustrations}

\begin{figure*}[t]
    \centering
    \includegraphics[width=1\textwidth]{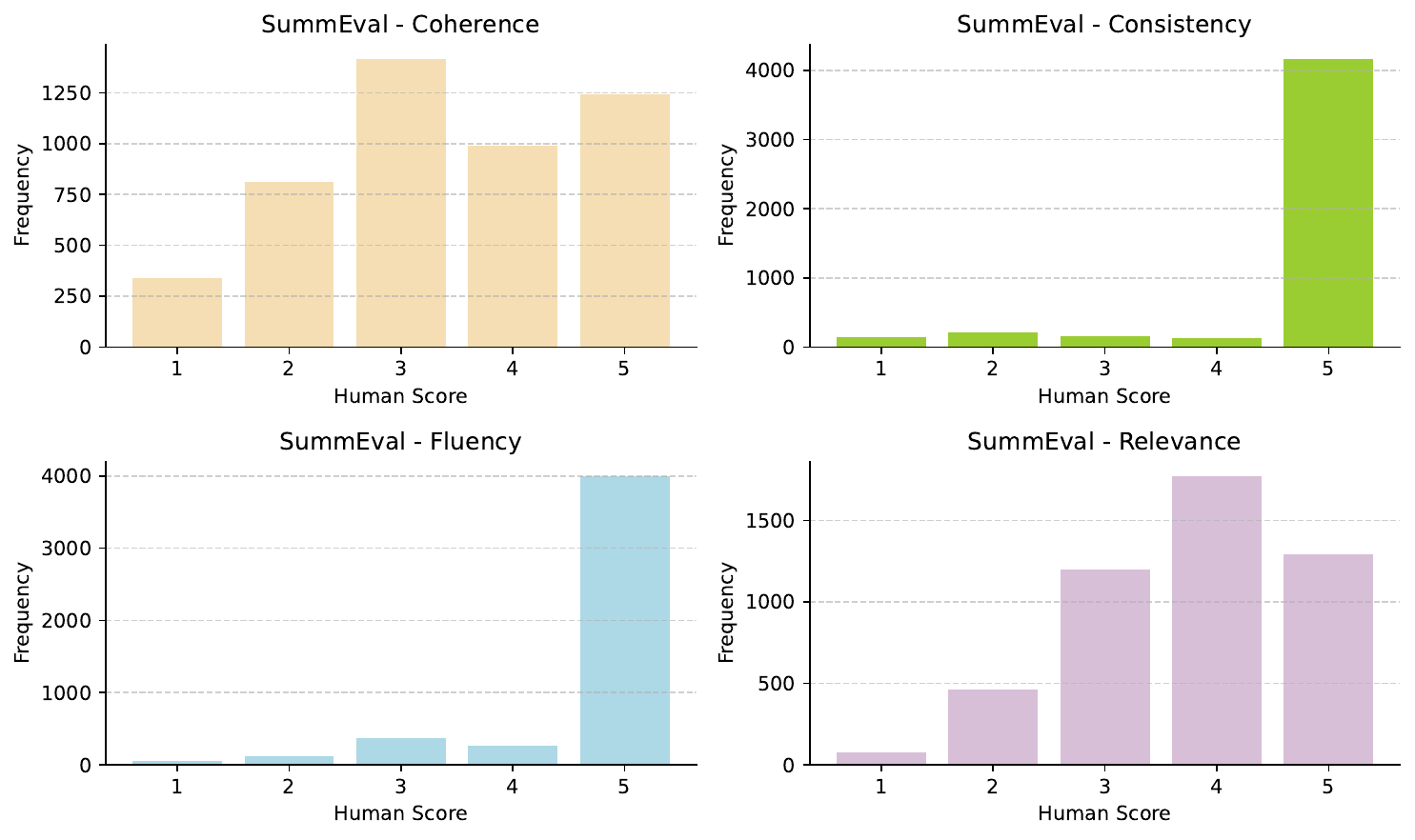}
    \caption{Illustration of the distribution of human evaluation scores of SummEval. The scores for the consistency and fluency aspects are highly skewed towards 5, which potentially leads to the decrease in performance of \hypoeval on theses aspects.}
    \label{fig:score_distribution}
\end{figure*}

To further show the exceptions discussed in \cref{sec:results_final}, we include the histograms of human annotation score distribution for the consistency and fluency aspects of SummEval in \cref{fig:score_distribution}.

\clearpage

\begin{thebibliography}{46}
\providecommand{\natexlab}[1]{#1}
\providecommand{\url}[1]{\texttt{#1}}
\expandafter\ifx\csname urlstyle\endcsname\relax
  \providecommand{\doi}[1]{doi: #1}\else
  \providecommand{\doi}{doi: \begingroup \urlstyle{rm}\Url}\fi

\bibitem[Bavaresco et~al.(2024)Bavaresco, Bernardi, Bertolazzi, Elliott, Fernández, Gatt, Ghaleb, Giulianelli, Hanna, Koller, Martins, Mondorf, Neplenbroek, Pezzelle, Plank, Schlangen, Suglia, Surikuchi, Takmaz, and Testoni]{bavaresco2024llmsinsteadhumanjudges}
Anna Bavaresco, Raffaella Bernardi, Leonardo Bertolazzi, Desmond Elliott, Raquel Fernández, Albert Gatt, Esam Ghaleb, Mario Giulianelli, Michael Hanna, Alexander Koller, André F.~T. Martins, Philipp Mondorf, Vera Neplenbroek, Sandro Pezzelle, Barbara Plank, David Schlangen, Alessandro Suglia, Aditya~K Surikuchi, Ece Takmaz, and Alberto Testoni.
\newblock Llms instead of human judges? a large scale empirical study across 20 nlp evaluation tasks, 2024.
\newblock URL \url{https://arxiv.org/abs/2406.18403}.

\bibitem[Chen et~al.(2023)Chen, Wang, Jiang, Shi, and Xu]{chen2023exploringuselargelanguage}
Yi~Chen, Rui Wang, Haiyun Jiang, Shuming Shi, and Ruifeng Xu.
\newblock Exploring the use of large language models for reference-free text quality evaluation: An empirical study, 2023.
\newblock URL \url{https://arxiv.org/abs/2304.00723}.

\bibitem[Chhun et~al.(2022)Chhun, Colombo, Clavel, and Suchanek]{chhun2022humancriteriaautomaticmetrics}
Cyril Chhun, Pierre Colombo, Chloé Clavel, and Fabian~M. Suchanek.
\newblock Of human criteria and automatic metrics: A benchmark of the evaluation of story generation, 2022.
\newblock URL \url{https://arxiv.org/abs/2208.11646}.

\bibitem[Chiang \& yi~Lee(2023)Chiang and yi~Lee]{chiang2023largelanguagemodelsalternative}
Cheng-Han Chiang and Hung yi~Lee.
\newblock Can large language models be an alternative to human evaluations?, 2023.
\newblock URL \url{https://arxiv.org/abs/2305.01937}.

\bibitem[Cook et~al.(2024)Cook, Rocktäschel, Foerster, Aumiller, and Wang]{cook2024tickingboxesgeneratedchecklists}
Jonathan Cook, Tim Rocktäschel, Jakob Foerster, Dennis Aumiller, and Alex Wang.
\newblock Ticking all the boxes: Generated checklists improve llm evaluation and generation, 2024.
\newblock URL \url{https://arxiv.org/abs/2410.03608}.

\bibitem[Dubey et~al.(2024)Dubey, Jauhri, Pandey, and et~al.]{dubey2024llama3herdmodels}
Abhimanyu Dubey, Abhinav Jauhri, Abhinav Pandey, and et~al.
\newblock The llama 3 herd of models, 2024.
\newblock URL \url{https://arxiv.org/abs/2407.21783}.

\bibitem[Fabbri et~al.(2021)Fabbri, Kryściński, McCann, Xiong, Socher, and Radev]{fabbri2021summevalreevaluatingsummarizationevaluation}
Alexander~R. Fabbri, Wojciech Kryściński, Bryan McCann, Caiming Xiong, Richard Socher, and Dragomir Radev.
\newblock Summeval: Re-evaluating summarization evaluation, 2021.
\newblock URL \url{https://arxiv.org/abs/2007.12626}.

\bibitem[Fang et~al.(2024)Fang, Liu, Kim, Bhedaru, Liu, Singh, Lipka, Mathur, Ahmed, Dernoncourt, Rossi, and Deilamsalehy]{fang2024multillmtextsummarization}
Jiangnan Fang, Cheng-Tse Liu, Jieun Kim, Yash Bhedaru, Ethan Liu, Nikhil Singh, Nedim Lipka, Puneet Mathur, Nesreen~K. Ahmed, Franck Dernoncourt, Ryan~A. Rossi, and Hanieh Deilamsalehy.
\newblock Multi-llm text summarization, 2024.
\newblock URL \url{https://arxiv.org/abs/2412.15487}.

\bibitem[Fu et~al.(2023)Fu, Ng, Jiang, and Liu]{fu2023gptscoreevaluatedesire}
Jinlan Fu, See-Kiong Ng, Zhengbao Jiang, and Pengfei Liu.
\newblock Gptscore: Evaluate as you desire, 2023.
\newblock URL \url{https://arxiv.org/abs/2302.04166}.

\bibitem[Grusky et~al.(2020)Grusky, Naaman, and Artzi]{grusky2020newsroomdataset13million}
Max Grusky, Mor Naaman, and Yoav Artzi.
\newblock Newsroom: A dataset of 1.3 million summaries with diverse extractive strategies, 2020.
\newblock URL \url{https://arxiv.org/abs/1804.11283}.

\bibitem[Gu et~al.(2025)Gu, Jiang, Shi, Tan, Zhai, Xu, Li, Shen, Ma, Liu, Wang, Zhang, Wang, Gao, Ni, and Guo]{gu2025surveyllmasajudge}
Jiawei Gu, Xuhui Jiang, Zhichao Shi, Hexiang Tan, Xuehao Zhai, Chengjin Xu, Wei Li, Yinghan Shen, Shengjie Ma, Honghao Liu, Saizhuo Wang, Kun Zhang, Yuanzhuo Wang, Wen Gao, Lionel Ni, and Jian Guo.
\newblock A survey on llm-as-a-judge, 2025.
\newblock URL \url{https://arxiv.org/abs/2411.15594}.

\bibitem[Krishna et~al.(2021)Krishna, Roy, and Iyyer]{krishna-etal-2021-hurdles}
Kalpesh Krishna, Aurko Roy, and Mohit Iyyer.
\newblock Hurdles to progress in long-form question answering.
\newblock In Kristina Toutanova, Anna Rumshisky, Luke Zettlemoyer, Dilek Hakkani-Tur, Iz~Beltagy, Steven Bethard, Ryan Cotterell, Tanmoy Chakraborty, and Yichao Zhou (eds.), \emph{Proceedings of the 2021 Conference of the North American Chapter of the Association for Computational Linguistics: Human Language Technologies}, pp.\  4940--4957, Online, June 2021. Association for Computational Linguistics.
\newblock \doi{10.18653/v1/2021.naacl-main.393}.
\newblock URL \url{https://aclanthology.org/2021.naacl-main.393/}.

\bibitem[Krumdick et~al.(2025)Krumdick, Lovering, Reddy, Ebner, and Tanner]{krumdick2025freelabelslimitationsllmasajudge}
Michael Krumdick, Charles Lovering, Varshini Reddy, Seth Ebner, and Chris Tanner.
\newblock No free labels: Limitations of llm-as-a-judge without human grounding, 2025.
\newblock URL \url{https://arxiv.org/abs/2503.05061}.

\bibitem[Lee et~al.(2024)Lee, Kim, Kim, Cho, and Kang]{lee2024checkevalrobustevaluationframework}
Yukyung Lee, Joonghoon Kim, Jaehee Kim, Hyowon Cho, and Pilsung Kang.
\newblock Checkeval: Robust evaluation framework using large language model via checklist, 2024.
\newblock URL \url{https://arxiv.org/abs/2403.18771}.

\bibitem[Li et~al.(2025)Li, Jiang, Huang, Beigi, Zhao, Tan, Bhattacharjee, Jiang, Chen, Wu, Shu, Cheng, and Liu]{li2025generationjudgmentopportunitieschallenges}
Dawei Li, Bohan Jiang, Liangjie Huang, Alimohammad Beigi, Chengshuai Zhao, Zhen Tan, Amrita Bhattacharjee, Yuxuan Jiang, Canyu Chen, Tianhao Wu, Kai Shu, Lu~Cheng, and Huan Liu.
\newblock From generation to judgment: Opportunities and challenges of llm-as-a-judge, 2025.
\newblock URL \url{https://arxiv.org/abs/2411.16594}.

\bibitem[Li et~al.(2024{\natexlab{a}})Li, Chen, Ai, Chu, Zhou, Dong, and Liu]{li2024calibraevalcalibratingpredictiondistribution}
Haitao Li, Junjie Chen, Qingyao Ai, Zhumin Chu, Yujia Zhou, Qian Dong, and Yiqun Liu.
\newblock Calibraeval: Calibrating prediction distribution to mitigate selection bias in llms-as-judges, 2024{\natexlab{a}}.
\newblock URL \url{https://arxiv.org/abs/2410.15393}.

\bibitem[Li et~al.(2024{\natexlab{b}})Li, Wang, Ma, Wu, Wang, Gao, and Liu]{li2024splitmergealigningposition}
Zongjie Li, Chaozheng Wang, Pingchuan Ma, Daoyuan Wu, Shuai Wang, Cuiyun Gao, and Yang Liu.
\newblock Split and merge: Aligning position biases in llm-based evaluators, 2024{\natexlab{b}}.
\newblock URL \url{https://arxiv.org/abs/2310.01432}.

\bibitem[Lin(2004)]{lin2004rouge}
Chin-Yew Lin.
\newblock Rouge: A package for automatic evaluation of summaries.
\newblock In \emph{Text summarization branches out}, pp.\  74--81, 2004.

\bibitem[Liu et~al.(2025{\natexlab{a}})Liu, Zhou, Li, Yuan, and Tan]{liu2025literaturemeetsdatasynergistic}
Haokun Liu, Yangqiaoyu Zhou, Mingxuan Li, Chenfei Yuan, and Chenhao Tan.
\newblock Literature meets data: A synergistic approach to hypothesis generation, 2025{\natexlab{a}}.
\newblock URL \url{https://arxiv.org/abs/2410.17309}.

\bibitem[Liu et~al.(2024{\natexlab{a}})Liu, Shen, Xu, Cao, Cho, Kumar, Ghanadan, and Huang]{liu-etal-2024-x}
Minqian Liu, Ying Shen, Zhiyang Xu, Yixin Cao, Eunah Cho, Vaibhav Kumar, Reza Ghanadan, and Lifu Huang.
\newblock {X}-eval: Generalizable multi-aspect text evaluation via augmented instruction tuning with auxiliary evaluation aspects.
\newblock In Kevin Duh, Helena Gomez, and Steven Bethard (eds.), \emph{Proceedings of the 2024 Conference of the North American Chapter of the Association for Computational Linguistics: Human Language Technologies (Volume 1: Long Papers)}, pp.\  8560--8579, Mexico City, Mexico, June 2024{\natexlab{a}}. Association for Computational Linguistics.
\newblock \doi{10.18653/v1/2024.naacl-long.473}.
\newblock URL \url{https://aclanthology.org/2024.naacl-long.473/}.

\bibitem[Liu \& Chen(2016)Liu and Chen]{NIPS2016_d79aac07}
Yang Liu and Yiling Chen.
\newblock A bandit framework for strategic regression.
\newblock In D.~Lee, M.~Sugiyama, U.~Luxburg, I.~Guyon, and R.~Garnett (eds.), \emph{Advances in Neural Information Processing Systems}, volume~29. Curran Associates, Inc., 2016.
\newblock URL \url{https://proceedings.neurips.cc/paper_files/paper/2016/file/d79aac075930c83c2f1e369a511148fe-Paper.pdf}.

\bibitem[Liu et~al.(2023{\natexlab{a}})Liu, Iter, Xu, Wang, Xu, and Zhu]{liu2023gevalnlgevaluationusing}
Yang Liu, Dan Iter, Yichong Xu, Shuohang Wang, Ruochen Xu, and Chenguang Zhu.
\newblock G-eval: Nlg evaluation using gpt-4 with better human alignment, 2023{\natexlab{a}}.
\newblock URL \url{https://arxiv.org/abs/2303.16634}.

\bibitem[Liu et~al.(2025{\natexlab{b}})Liu, Zhou, Guo, Shareghi, Vulić, Korhonen, and Collier]{liu2025aligninghumanjudgementrole}
Yinhong Liu, Han Zhou, Zhijiang Guo, Ehsan Shareghi, Ivan Vulić, Anna Korhonen, and Nigel Collier.
\newblock Aligning with human judgement: The role of pairwise preference in large language model evaluators, 2025{\natexlab{b}}.
\newblock URL \url{https://arxiv.org/abs/2403.16950}.

\bibitem[Liu et~al.(2023{\natexlab{b}})Liu, Yang, Huang, Zhang, Huang, Wei, Deng, Sun, and Zhang]{liu2023calibratingllmbasedevaluator}
Yuxuan Liu, Tianchi Yang, Shaohan Huang, Zihan Zhang, Haizhen Huang, Furu Wei, Weiwei Deng, Feng Sun, and Qi~Zhang.
\newblock Calibrating llm-based evaluator, 2023{\natexlab{b}}.
\newblock URL \url{https://arxiv.org/abs/2309.13308}.

\bibitem[Liu et~al.(2024{\natexlab{b}})Liu, Yang, Huang, Zhang, Huang, Wei, Deng, Sun, and Zhang]{liu-etal-2024-calibrating}
Yuxuan Liu, Tianchi Yang, Shaohan Huang, Zihan Zhang, Haizhen Huang, Furu Wei, Weiwei Deng, Feng Sun, and Qi~Zhang.
\newblock Calibrating {LLM}-based evaluator.
\newblock In Nicoletta Calzolari, Min-Yen Kan, Veronique Hoste, Alessandro Lenci, Sakriani Sakti, and Nianwen Xue (eds.), \emph{Proceedings of the 2024 Joint International Conference on Computational Linguistics, Language Resources and Evaluation (LREC-COLING 2024)}, pp.\  2638--2656, Torino, Italia, May 2024{\natexlab{b}}. ELRA and ICCL.
\newblock URL \url{https://aclanthology.org/2024.lrec-main.237/}.

\bibitem[Liusie et~al.(2024)Liusie, Manakul, and Gales]{liusie2024llmcomparativeassessmentzeroshot}
Adian Liusie, Potsawee Manakul, and Mark J.~F. Gales.
\newblock Llm comparative assessment: Zero-shot nlg evaluation through pairwise comparisons using large language models, 2024.
\newblock URL \url{https://arxiv.org/abs/2307.07889}.

\bibitem[Lo et~al.(2020)Lo, Wang, Neumann, Kinney, and Weld]{lo-wang-2020-s2orc}
Kyle Lo, Lucy~Lu Wang, Mark Neumann, Rodney Kinney, and Daniel Weld.
\newblock {S}2{ORC}: The semantic scholar open research corpus.
\newblock In \emph{Proceedings of the 58th Annual Meeting of the Association for Computational Linguistics}, pp.\  4969--4983, Online, July 2020. Association for Computational Linguistics.
\newblock \doi{10.18653/v1/2020.acl-main.447}.
\newblock URL \url{https://www.aclweb.org/anthology/2020.acl-main.447}.

\bibitem[Min et~al.(2023)Min, Krishna, Lyu, Lewis, tau Yih, Koh, Iyyer, Zettlemoyer, and Hajishirzi]{min2023factscorefinegrainedatomicevaluation}
Sewon Min, Kalpesh Krishna, Xinxi Lyu, Mike Lewis, Wen tau Yih, Pang~Wei Koh, Mohit Iyyer, Luke Zettlemoyer, and Hannaneh Hajishirzi.
\newblock Factscore: Fine-grained atomic evaluation of factual precision in long form text generation, 2023.
\newblock URL \url{https://arxiv.org/abs/2305.14251}.

\bibitem[OpenAI(2023)]{openai2023gpt4}
OpenAI.
\newblock {GPT}-4 technical report, 2023.

\bibitem[Papineni et~al.(2002)Papineni, Roukos, Ward, and Zhu]{papineni2002bleu}
Kishore Papineni, Salim Roukos, Todd Ward, and Wei-Jing Zhu.
\newblock Bleu: a method for automatic evaluation of machine translation.
\newblock In \emph{Proceedings of the 40th annual meeting of the Association for Computational Linguistics}, pp.\  311--318, 2002.

\bibitem[Pereira et~al.(2024)Pereira, Assumpcao, and Lotufo]{pereira2024checkevalchecklistbasedapproachevaluating}
Jayr Pereira, Andre Assumpcao, and Roberto Lotufo.
\newblock Check-eval: A checklist-based approach for evaluating text quality, 2024.
\newblock URL \url{https://arxiv.org/abs/2407.14467}.

\bibitem[Qin et~al.(2024{\natexlab{a}})Qin, Song, Hu, Yao, Cho, Wang, Wu, Liu, Liu, and Yu]{qin2024infobenchevaluatinginstructionfollowing}
Yiwei Qin, Kaiqiang Song, Yebowen Hu, Wenlin Yao, Sangwoo Cho, Xiaoyang Wang, Xuansheng Wu, Fei Liu, Pengfei Liu, and Dong Yu.
\newblock Infobench: Evaluating instruction following ability in large language models, 2024{\natexlab{a}}.
\newblock URL \url{https://arxiv.org/abs/2401.03601}.

\bibitem[Qin et~al.(2024{\natexlab{b}})Qin, Jagerman, Hui, Zhuang, Wu, Yan, Shen, Liu, Liu, Metzler, Wang, and Bendersky]{qin2024largelanguagemodelseffective}
Zhen Qin, Rolf Jagerman, Kai Hui, Honglei Zhuang, Junru Wu, Le~Yan, Jiaming Shen, Tianqi Liu, Jialu Liu, Donald Metzler, Xuanhui Wang, and Michael Bendersky.
\newblock Large language models are effective text rankers with pairwise ranking prompting, 2024{\natexlab{b}}.
\newblock URL \url{https://arxiv.org/abs/2306.17563}.

\bibitem[Que et~al.(2024)Que, Duan, He, Mou, Zhou, Liu, Rong, Wang, Yang, Zhang, Peng, Zhang, Zhang, and Chen]{que2024hellobenchevaluatinglongtext}
Haoran Que, Feiyu Duan, Liqun He, Yutao Mou, Wangchunshu Zhou, Jiaheng Liu, Wenge Rong, Zekun~Moore Wang, Jian Yang, Ge~Zhang, Junran Peng, Zhaoxiang Zhang, Songyang Zhang, and Kai Chen.
\newblock Hellobench: Evaluating long text generation capabilities of large language models, 2024.
\newblock URL \url{https://arxiv.org/abs/2409.16191}.

\bibitem[Sclar et~al.(2024)Sclar, Choi, Tsvetkov, and Suhr]{sclar2024quantifyinglanguagemodelssensitivity}
Melanie Sclar, Yejin Choi, Yulia Tsvetkov, and Alane Suhr.
\newblock Quantifying language models' sensitivity to spurious features in prompt design or: How i learned to start worrying about prompt formatting, 2024.
\newblock URL \url{https://arxiv.org/abs/2310.11324}.

\bibitem[Tan et~al.(2024)Tan, Guo, Shi, Xu, Liu, Feng, Li, Wang, Shang, Liu, and Song]{tan2024proxyqaalternativeframeworkevaluating}
Haochen Tan, Zhijiang Guo, Zhan Shi, Lu~Xu, Zhili Liu, Yunlong Feng, Xiaoguang Li, Yasheng Wang, Lifeng Shang, Qun Liu, and Linqi Song.
\newblock Proxyqa: An alternative framework for evaluating long-form text generation with large language models, 2024.
\newblock URL \url{https://arxiv.org/abs/2401.15042}.

\bibitem[Wei et~al.(2022)Wei, Wang, Schuurmans, Bosma, ichter, Xia, Chi, Le, and Zhou]{NEURIPS2022_9d560961}
Jason Wei, Xuezhi Wang, Dale Schuurmans, Maarten Bosma, brian ichter, Fei Xia, Ed~Chi, Quoc~V Le, and Denny Zhou.
\newblock Chain-of-thought prompting elicits reasoning in large language models.
\newblock In S.~Koyejo, S.~Mohamed, A.~Agarwal, D.~Belgrave, K.~Cho, and A.~Oh (eds.), \emph{Advances in Neural Information Processing Systems}, volume~35, pp.\  24824--24837. Curran Associates, Inc., 2022.
\newblock URL \url{https://proceedings.neurips.cc/paper_files/paper/2022/file/9d5609613524ecf4f15af0f7b31abca4-Paper-Conference.pdf}.

\bibitem[xAI(2025)]{xai2025grok3}
xAI.
\newblock Grok 3 beta — the age of reasoning agents.
\newblock \url{https://x.ai/news/grok-3}, February 2025.
\newblock Accessed: 2025-03-27.

\bibitem[Yao et~al.(2024)Yao, Jiang, Bobinac, Yang, and Hu]{yao2024benchmarkingmachinetranslationcultural}
Binwei Yao, Ming Jiang, Tara Bobinac, Diyi Yang, and Junjie Hu.
\newblock Benchmarking machine translation with cultural awareness, 2024.
\newblock URL \url{https://arxiv.org/abs/2305.14328}.

\bibitem[Yuan et~al.(2021)Yuan, Neubig, and Liu]{yuan2021bartscoreevaluatinggeneratedtext}
Weizhe Yuan, Graham Neubig, and Pengfei Liu.
\newblock Bartscore: Evaluating generated text as text generation, 2021.
\newblock URL \url{https://arxiv.org/abs/2106.11520}.

\bibitem[Yue et~al.(2023)Yue, Wang, Chen, Zhang, Su, and Sun]{yue2023automaticevaluationattributionlarge}
Xiang Yue, Boshi Wang, Ziru Chen, Kai Zhang, Yu~Su, and Huan Sun.
\newblock Automatic evaluation of attribution by large language models, 2023.
\newblock URL \url{https://arxiv.org/abs/2305.06311}.

\bibitem[Zhang et~al.(2020)Zhang, Kishore, Wu, Weinberger, and Artzi]{zhang2020bertscoreevaluatingtextgeneration}
Tianyi Zhang, Varsha Kishore, Felix Wu, Kilian~Q. Weinberger, and Yoav Artzi.
\newblock Bertscore: Evaluating text generation with bert, 2020.
\newblock URL \url{https://arxiv.org/abs/1904.09675}.

\bibitem[Zhong et~al.(2022)Zhong, Liu, Yin, Mao, Jiao, Liu, Zhu, Ji, and Han]{zhong2022unifiedmultidimensionalevaluatortext}
Ming Zhong, Yang Liu, Da~Yin, Yuning Mao, Yizhu Jiao, Pengfei Liu, Chenguang Zhu, Heng Ji, and Jiawei Han.
\newblock Towards a unified multi-dimensional evaluator for text generation, 2022.
\newblock URL \url{https://arxiv.org/abs/2210.07197}.

\bibitem[Zhou et~al.(2024{\natexlab{a}})Zhou, Wan, Liu, Collier, Vulić, and Korhonen]{zhou2024fairerpreferenceselicitimproved}
Han Zhou, Xingchen Wan, Yinhong Liu, Nigel Collier, Ivan Vulić, and Anna Korhonen.
\newblock Fairer preferences elicit improved human-aligned large language model judgments, 2024{\natexlab{a}}.
\newblock URL \url{https://arxiv.org/abs/2406.11370}.

\bibitem[Zhou et~al.(2023)Zhou, Lu, Mishra, Brahma, Basu, Luan, Zhou, and Hou]{zhou2023instructionfollowingevaluationlargelanguage}
Jeffrey Zhou, Tianjian Lu, Swaroop Mishra, Siddhartha Brahma, Sujoy Basu, Yi~Luan, Denny Zhou, and Le~Hou.
\newblock Instruction-following evaluation for large language models, 2023.
\newblock URL \url{https://arxiv.org/abs/2311.07911}.

\bibitem[Zhou et~al.(2024{\natexlab{b}})Zhou, Liu, Srivastava, Mei, and Tan]{Zhou_2024}
Yangqiaoyu Zhou, Haokun Liu, Tejes Srivastava, Hongyuan Mei, and Chenhao Tan.
\newblock Hypothesis generation with large language models.
\newblock In \emph{Proceedings of the 1st Workshop on NLP for Science (NLP4Science)}, pp.\  117–139. Association for Computational Linguistics, 2024{\natexlab{b}}.
\newblock \doi{10.18653/v1/2024.nlp4science-1.10}.
\newblock URL \url{http://dx.doi.org/10.18653/v1/2024.nlp4science-1.10}.

\end{thebibliography}
\end{document}